\documentclass[journal]{IEEEtran}
\usepackage{cite}
\usepackage{amsmath,amssymb,amsfonts}
\usepackage{algorithmic}
\usepackage[ruled,vlined,linesnumbered]{algorithm2e}
\SetAlCapNameFnt{\footnotesize}
\SetAlCapFnt{\footnotesize}
\SetAlgorithmName{Algorithm}{Algorithm}{List of Algorithms}
\usepackage{graphicx}
\usepackage{stmaryrd}
\usepackage{xcolor}
\usepackage{array}
\usepackage{commath}
\usepackage{footnote}
\usepackage{sidecap}
\usepackage{stfloats}
\usepackage{tabularx, boldline}
\usepackage{rotating,booktabs,multirow}
\usepackage{flexisym}
\usepackage{breqn}
\usepackage{url}
\usepackage{adjustbox}
\usepackage{makecell}
\usepackage{textcomp}
\usepackage{amsmath}
\usepackage{graphicx}%,subcaption}
\usepackage{acronym}
\usepackage{tabularx, boldline}
\usepackage{rotating,booktabs,multirow}
\usepackage{enumerate}
\usepackage{mathtools}
\usepackage{dsfont}
\usepackage{nomencl}
\makenomenclature

\def\E{\mathbb{E}} % Esperanza
\newtheorem{proposition}{\it Proposition}
\usepackage{authblk}
\def\BibTeX{{\rm B\kern-.05em{\sc i\kern-.025em b}\kern-.08em
    T\kern-.1667em\lower.7ex\hbox{E}\kern-.125emX}}

\def\P{\mathbb{P}}

\usepackage[ruled]{algorithm2e}
\usepackage[hang,flushmargin]{footmisc}
\usepackage{lipsum}
\makeatletter
\newcommand{\algorithmfootnote}[2][\footnotesize]{%
  \let\old@algocf@finish\@algocf@finish% Store algorithm finish macro
  \def\@algocf@finish{\old@algocf@finish% Update finish macro to insert "footnote"
    \leavevmode\rlap{\begin{minipage}{\linewidth}
    #1#2
    \end{minipage}}%
  }%
}
\makeatother

\ifCLASSINFOpdf

\else

\fi

\begin{document}


%\title{Privacy-Cost Energy Management in Smart Meters: Mutual Information Based Privacy}
%\title{On the Impact of the Privacy Measure on Smart Meters Privacy-Cost Energy Management}
	
\title{Preserving Privacy in GANs Against Membership Inference Attack}

\author{Mohammadhadi~Shateri,~\IEEEmembership{Member,~IEEE,} Francisco~Messina,
  Fabrice~Labeau,~\IEEEmembership{Senior Member,~IEEE, }%
        Pablo~Piantanida,~\IEEEmembership{Senior Member,~IEEE}

\thanks{M. Shateri is with the Department of Systems Engineering, École de technologie supérieure, QC, Canada. (e-mail: mohammadhadi.shateri@etsmtl.ca).}
% <-this % stops a space
\thanks{F. Messina is with the School of Engineering - Universidad de Buenos Aires, Buenos Aires, Argentina (e-mail: fmessina@fi.uba.ar).}% <-this % stops a space
\thanks{F. Labeau are with the Department of Electrical and Computer Engineering, McGill University, QC, Canada. (e-mail: fabrice.labeau@mcgill.ca).}% <-this % stops a space 
\thanks{P. Piantanida is with ILLS - International Laboratory on Learning Systems and MILA - Quebec AI Institute; CNRS, CentraleSupélec; Montreal, QC H3C 1K3, Canada.  (e-mail: pablo.piantanida@cnrs.fr).}

%\thanks{F. Labeau  is with the Department of Electrical and Computer Engineering, McGill University, QC, Canada (e-mail: fabrice.labeau@mcgill.ca).}% <-this % stops a space
}
\maketitle

\IEEEpeerreviewmaketitle

\begin{abstract}
Generative Adversarial Networks (GANs) have been widely used for generating synthetic data for cases where there is a limited size real-world data set or when data holders are unwilling to share their data samples. Recent works showed that GANs, due to overfitting and memorization, might leak information regarding their training data samples. This makes GANs vulnerable to Membership Inference Attacks (MIAs). Several defense strategies have been proposed in the literature to mitigate this privacy issue. Unfortunately, defense strategies based on differential privacy are proven to reduce extensively the quality of the synthetic data points. On the other hand, more recent frameworks such as PrivGAN and PAR-GAN are not suitable for small-size training data sets. In the present work, the overfitting in GANs is studied in terms of the discriminator, and a more general measure of overfitting based on the Bhattacharyya coefficient is defined. Then, inspired by Fano's inequality, our first defense mechanism against MIAs is proposed. This framework, which requires only a simple modification in the loss function of GANs, is referred to as the maximum entropy GAN or MEGAN and significantly improves the robustness of GANs to MIAs. As a second defense strategy, a more heuristic model based on minimizing the information leaked from the generated samples about the training data points is presented. This approach is referred to as mutual information minimization GAN (MIMGAN) and uses a variational representation of the mutual information to minimize the information that a synthetic sample might leak about the whole training data set. Applying the proposed frameworks to some commonly used data sets against state-of-the-art MIAs reveals that the proposed methods can reduce the accuracy of the adversaries to the level of random guessing accuracy with a small reduction in the quality of the synthetic data samples.

%\francisco{Abstract is a little bit long, around 309 words now. It would probably be good and maybe necessary to reduce it to 200-250 words (we need to check the journal requirements).}

\end{abstract}


%\nomenclature[1]{$H(X)$}{Entropy of random variable $X$}
%\nomenclature[2]{$I(X^T \to Y^T)$}{Directed information between $X^T$ and $Y^T$}
%\nomenclature[3]{${X \mkv Y \mkv Z}$}{Markov chain among $X$, $Y$ and $Z$}
%\printnomenclature

\section{Introduction}

\subsection{Motivation}
Recent advances in the development of novel algorithms in machine learning and data analysis are mostly due to the availability of publicly accessible data sets and the possibility of data sharing. At the same time, several concerns have been raised regarding the violation of users' privacy, since adversaries can infer sensitive information about individuals by analyzing the open access data sets~\cite{primault2018long,shateri2020real,8611220}. This is one of the main reasons why in some fields (e.g., medical applications, power systems, finances) there is a shortage of public real-world data sets. A promising solution to this problem is the development of data generators/synthesizers for producing synthetic data samples from the same underlying distribution of a given sensitive real-world data set~\cite{niu2020defect,9034117, 9372157} while avoiding sharing data directly. 

Generative adversarial networks (GANs) introduced firstly by Goodfellow et. al~\cite{goodfellow2014generative} have been extensively used for generating synthetic data in applications where there is limited access to real-world data sets. Despite their undeniable benefits, GANs are prone to overfitting and memorization of their training data sets~\cite{DBLP:conf/iclr/BrockDS19,zhao2020differentiable}. This makes them vulnerable to several privacy attacks such as membership inference attacks (MIAs)~\cite{hayes2019logan,hilprecht2019monte,liu2019performing,Chen2019GANLeaksAT}.  To be more specific, if a GAN experiences overfitting or memorizes the training data, it subsequently facilitates an adversary's ability to differentiate between generated samples that originate from the training dataset and those that do not. Furthermore, its discriminator assigns elevated scores to the data points that were included in the training set. This implies that the influence of overfitting and memorization on the accuracy of Membership Inference Attacks (MIA) is substantial.  This point is discussed in more detail in subsections~\ref{ove_mem}\& \ref{mia_in}. As a consequence, there is a high practical interest in developing privacy-aware training mechanisms for GANs.

\subsection{Related work}

MIAs on machine learning models were proposed for the first time by Shokri et. al in~\cite{shokri2017membership}. Recently, many different MIA strategies have been proposed that are effective on the classification/regression models~\cite{yeom2018privacy,nasr2019comprehensive,leino2020stolen,song2021systematic,choquette2021label}. However, developing MIAs on generative models is more challenging and the strategies studied for classification/regression models typically do not perform well. For example, in~\cite{hayes2019logan} an MIA based on shadow training, inspired by the MIA in~\cite{shokri2017membership}, was applied to a GAN trained with LFW face data sets, and the adversary performance was found to be similar to random guessing.

Membership inference attacks on GAN models can be generally classified into two categories based on their target, either focusing on the generator or the discriminator, namely, Discriminator-based Membership Inference Attacks (DMIA) and Generator-based Membership Inference Attacks (GMIA).

The first DMIAs were introduced in~\cite{hayes2019logan} where the adversary's goal is to distinguish data points used in the training dataset, accomplished by accessing the target GAN's discriminator. Empirical results on various datasets reveal the effectiveness of this strategy, achieving very high accuracy, even 100\% in some cases. In addition, Mukherjee et al.~\cite{mukherjee2021privgan} introduced an upper limit for membership inference accuracy in white-box MIAs on a GAN's discriminator, determined by their Total Variation Distance (TVD) attack, which estimates the total variation distance between the distribution of discriminator scores on the training and holdout datasets.

Shifting to GMIAs, a Monte Carlo (MC) attack was introduced in~\cite{hilprecht2019monte}. This MIA leverages Monte Carlo integration to exploit synthetic samples that are in the vicinity of the target sample, inferring the likelihood that the target sample belongs to the training dataset. This approach particularly excels in set membership inference, where the adversary determines whether a set of data points belongs to the training dataset or not. A co-membership inference attack strategy, akin to the MC approach but using the L$_2$-distance metric, was proposed in~\cite{liu2019performing}. However, this method is computationally more complex than existing approaches. The first taxonomy of MIAs on generative models was outlined in~\cite{Chen2019GANLeaksAT}, introducing a generic attack model in which the adversary aims to reconstruct the closest synthetic data point to a target sample. This approach relies on the generator's ability to produce synthetic samples resembling the training set, with the distance between the reconstructed synthetic sample and the target sample used to calculate the probability that the target sample belongs to the training data set. Extensive empirical comparisons across different scenarios and datasets have demonstrated the efficiency of this MIA strategy compared to other MIAs against generative models~\cite{Chen2019GANLeaksAT}. Another recent MIA targeting the generator of GANs was developed in~\cite{9720436}, where an auto-encoder, with the target generator as its decoder, is trained based on generated samples and their associated latent samples. During inference, the reconstruction error of a target sample, processed through the trained auto-encoder, is used to infer its membership label. This attack proved to be successful in cases when the number of training samples was small while not much better than a random attacker in other cases.

Taking into account all the proposed MIAs on GANs, it is widely acknowledged that the white-box attacks targeting the GAN's discriminator (DMIAs), specifically the TVD-based attack~\cite{mukherjee2021privgan} and the MIA introduced in reference~\cite{hayes2019logan}, are among the most effective MIAs against GANs. In contrast, other attack methods do not exhibit significantly improved performance compared to random attacks. Consequently, in our study, we assess the effectiveness of the proposed defense mechanisms based on the MIA strategies outlined in~\cite{hayes2019logan,mukherjee2021privgan}.

To address the vulnerability of generative models to MIAs, several defenses have been presented in the literature. The main idea of most of these frameworks is based on differential privacy (DP)~\cite{jordon2018pate,xie2018differentially,triastcyn2019generating,wu2019generalization,xu2019ganobfuscator}. Although these frameworks were shown effective in preventing membership inference by adversaries, DP-based GANs degrade significantly the quality of synthetic samples~\cite{hayes2019logan,chen2018differentially}. For more details on DP-GANs, the reader is referred to~\cite{fan2020survey}. On the other hand, a regularization technique known as dropout has been suggested in~\cite{hayes2019logan,hilprecht2019monte} in order to improve generalization in generative models and mitigate the membership inference issue. Using dropout techniques in GANs raises two main issues: determining the optimal dropout rate and placement, often requiring trial and error, and the significant slowdown in training, which can be challenging due to GANs' inherent instability. More sophisticated defenses designed specifically against MIAs in GANs were presented recently, including PrivGan~\cite{mukherjee2021privgan} and PAR-GAN~\cite{chen2021gan}. In both methods, the training data set is split into $N$ disjoint sub-sets. In PrivGan, a GAN (including a generator and a discriminator) is trained for each sub-set and the generators are trained to not only fool their associated discriminator but also prevent a classifier from distinguishing their generated samples from the other generators' samples. On the other hand, in the PAR-GAN, a single generator is trained to fight with $N$ discriminators (associated with $N$ disjoint data subsets). Considering the empirical results of these methods applied to several data sets, although both approaches showed to be effective (particularly for large values of $N$) in improving the generalization and mitigating the information leakage exploited by MIAs, they increase the computational complexity of the GAN training procedure quite significantly without providing any mathematical guarantees for reducing the overfitting or memorization. In addition, due to the inherent requirement of these models to use data partitions, they are not appropriate for cases in which there is a limited-size training data set. It also should be noted that for the PrivGan which includes $N$ generators, although a random selection strategy is suggested in\cite{mukherjee2021privgan}, it is not clear what is the best approach for sharing the final synthetic data samples.  

In this work, we adopt measures from information theory and statistics to modify the GAN framework for the sake of making it robust to MIA. To gain a deeper understanding of overfitting in GANs, we utilize the Bhattacharyya coefficient, calculated at the discriminator's output, as a metric for overfitting assessment. Subsequently, we introduce a defense mechanism that addresses this measure of overfitting, named the Maximum Entropy GAN (MEGAN). MEGAN is a straightforward modification of the conventional GAN that ensures robust learning of training data distributions while reducing MIA accuracy to a level akin to random attacks. Additionally, our study puts forth a second heuristic defense approach, the Mutual Information Minimization GAN (MIMGAN), which minimizes the mutual information between generated and training data through a variational representation, offering a practical strategy for guarding against MIAs. Experimental studies are done using four commonly used datasets including MNIST, fashion-MNIST, Chest X-ray images (Pneumonia), and Anime Faces datasets. The performance of the proposed defense mechanisms is evaluated compared with state-of-the-art models such as PrivGAN and DP-GAN, in terms of their robustness against MIAs and the fidelity and diversity of generated samples.

\subsection{Contributions}

The main contributions of this work can be summarized as follows: 
\begin{itemize}
    \item We study overfitting in GANs in terms of the discriminator by using the Bhattacharyya coefficient and discuss its relation and advantage over the classical notion of overfitting (i.e., generalization gap).
    \item Considering this new perspective of overfitting and the well-known Fano's inequality applied to the discriminator error, we propose the maximum entropy GAN (MEGAN) as a defense mechanism to state-of-the-art MIAs. MEGAN is a simple modification of the vanilla GAN that ensures learning the distribution of training data and at the same time reducing the MIAs accuracy to a level similar to that of a random attacker.
    \item As a second defense framework, a heuristic model based on minimizing the mutual information between the generated data and training data is proposed. In order to provide a simple implementation of this idea, we consider a variational representation of the mutual information. This method is referred to as the mutual information minimization GAN (MIMGAN) in the following. 
    %\item We perform an extensive statistical study based on the most popular data sets used in the field to evaluate the performance of the proposed frameworks in terms of robustness to MIAs, overfitting, memorization, and precision-recall of the generated data samples. %These results show that both MEGAN and MIMGAN can reduce the MIA inference towards the accuracy of a random attacker.
\end{itemize}

%\francisco{It may be good to add in contribution (iv) a summary of the final findings and comparison between the two proposed frameworks.}

\subsubsection*{Organization of the paper} The rest of the paper is organized as follows. In Section~\ref{sec:back} we present a background on GANs, the notions of overfitting and memorization in GANs, and MIAs on GANs. Then, in Section~\ref{sec:prop_fr}, the state-of-the-art in MIAs on GANs is considered from a new perspective and two defense mechanisms, MEGAN and MIMGAN, are proposed and discussed in detail. Extensive experimental results are presented and discussed in Section~\ref{sec:results}. Finally, concluding remarks are given in Section~\ref{sec:conclusion}.

\subsubsection*{Notation and mathematical conventions} Throughout this article, we use bold-face small letters, e.g. $\mathbf{x}$ to denote random variables, regular-face small letters, e.g. $x$ to denote specific realizations, and the capital letters, e.g. $X$ to refer to the set of values. $\mathbb{P}$ refers to the probability measure. $p$ is either a probability density function or a probability mass function. We use $p(x|y)$ to denote the conditional probability density function of $\mathbf{x}$ given $\mathbf{y}$. Only when it is necessary to avoid confusion we include a subscript in $p$. $\E [\cdot]$ is the expectation with respect to the joint distribution of all random variables involved; $H(\mathbf{x})$ is the Shannon entropy of random variable $\mathbf{x}$ and $I(\mathbf{x};\mathbf{y})$ is the mutual information between $\mathbf{x}$ and $\mathbf{y}$.

\section{Background}\label{sec:back}

\subsection{Generative Adversarial Networks}

The generative adversarial network (GAN) is comprised of two deep neural networks: a generator $G(\mathbf{z};\theta_g)$, with parameters $\theta_g$, and a discriminator $D(\mathbf{x};\theta_d)$, with parameters $\theta_d$, that are trained by competing in a minimax game. The generator receives random samples from the latent space (noise data $\mathbf{z}\sim p_{\mathbf{z}}$) as the input and aims to generate synthetic data by learning $p_{\mathbf{x_{tr}}}$, the probability distribution of the training dataset $X_{tr}$. On the other hand, the discriminator aims to determine the authenticity of the training data points versus the synthetic data points, by producing a real-valued number in the range $[0,1]$ interpreted as the probability that its input comes from original data rather than from the generator. The GAN framework is trained by playing the following minimax game with the value function $V(G,D)$~\cite{goodfellow2014generative}:
\begin{align}\label{objective_vanillaGAN}
    \underset{G}{\text{min}}\;\underset{D}{\text{max}}\; V(G,D)\; & =
    \E\left[\log D(\mathbf{x_{tr}})\right] +\nonumber\\
    &\E\left[\log(1-D(G(\mathbf{z})))\right],
\end{align}
where the training data points are modeled as independent and identically distributed (i.i.d.) samples of the random variable $\mathbf{x_{tr}}\sim p_{\mathbf{x_{tr}}}$. It can be shown that minimizing the GAN loss in equation~\eqref{objective_vanillaGAN} results in $p_{\mathbf{x_{g}}} = p_{\mathbf{x_{tr}}}$, i.e. the generator distribution would be identical to the distribution of the training dataset~\cite{goodfellow2014generative}. After training the GAN framework, the generator can be used for generating synthetic samples resembling real training data samples.

%\begin{figure}[htbp!]
%    \centering
%    \includegraphics[width=0.90\linewidth]{GAN_Framework.png}
%    \caption{General framework of a generative adversarial network (GAN). Both the generator and discriminator are modeled as deep neural networks.}
%    \label{fig:VGAN}
%\end{figure}

\subsection{Overfitting and Memorization in GANs}\label{ove_mem}
In discriminative models (classification and regression), overfitting is defined in terms of the difference between the performance of a model on the training data set and its performance on the test/hold-out data samples. In other words, a model overfits if it performs significantly better on the training data compared with the test data. However, for generative models, there is no generally accepted notion of overfitting. In the literature, overfitting in GANs is typically defined in terms of the discriminator output for the training and test data samples~\cite{wu2019generalization,mukherjee2021privgan,chen2021gan,brock2018large,yazici2020empirical,goodfellow2014generative}. This sometimes is referred to as the generalization gap and is defined as follows~\cite{chen2021gan}: 
\begin{equation}\label{eq:gengap}
    g_{\text{GAN}}^{D} = \E\left[\phi\left( D(\mathbf{x_{tr}})\right)\right] - \E\left[\phi\left( D(\mathbf{x_{te}})\right)\right],
\end{equation}
where $g_{\text{GAN}}^{D}$ is the generalization gap and $\phi(x)$ is selected as $\log(x)$ or simply as $x$.

Regarding memorization, the nearest neighbor test has been used extensively in the literature~\cite{huang2018empirical,meehan2020non,brock2018large,karras2018progressive,yazici2020empirical,goodfellow2014generative}, especially for the cases where a global sense of memorization is of interest~\cite{van2021memorization}. More specifically, considering the premise that memorization in GANs occurs when the generated data samples are closer to the training dataset than the actual samples from the test set, equation~\eqref{eq:memgan} is used:
\begin{equation}\label{eq:memgan}
    m_{\text{GAN}}^{G} = \frac{\E\left[\underset{x\in X_{TR}}{\text{min }} d\left(\mathbf{x_{te}}, x\right)\right]}{\E\left[\underset{x\in X_{TR}}{\text{min }} d\left(G(\mathbf{z}), x\right)\right]},
\end{equation}
where $m_{\text{GAN}}^{G}$ is the measure of memorization in GANs, and $d(\cdot, \cdot)$ is a distance function (e.g., the Euclidean distance). It should be noted that $m_{\text{GAN}}^{G}>1$ can be interpreted as memorization. %It should be noted that for the notation of memorization at an instance level, especially for explicit density models such as variational autoencoders, the reader is referred to~\cite{van2021memorization} and this is out of the scope of our study.

%It should be noted that for $g_{\text{GAN}}^{D}$ very close to zero and $m_{\text{GAN}}^{G}$ less than one.

\subsection{Membership Inference Attack}\label{mia_in}
In a membership inference attack (MIA) to a GAN, the goal of the attacker is to determine whether a specific data sample $x\in X$ was used in the training of the target GAN or not. In these attacks, an adversary aims to infer whether an instance or a set of target samples was used to train a specific model or not. Depending on the information available to the adversary, the MIAs are categorized into two main families: black-box attacks and white-box attacks. In the former, it is assumed that the adversary can only get (unlimited) query access to the target model, while in the latter the adversary has full knowledge of the model parameters. In most MIA frameworks on GANs, it is assumed that the attacker has access to a data pool $X$ that includes the training data set (i.e., $X_{tr}\subset X$) without having knowledge about training data samples. Therefore, an attacker seeks to learn a mapping $M_{\text{GAN}}: X \longrightarrow [0,1]$, where $M_{\text{GAN}}(x)$ can be interpreted as the likelihood of $x \in X_{tr}$.

\begin{figure}[htbp!]
    \centering
    \includegraphics[width=1\linewidth]{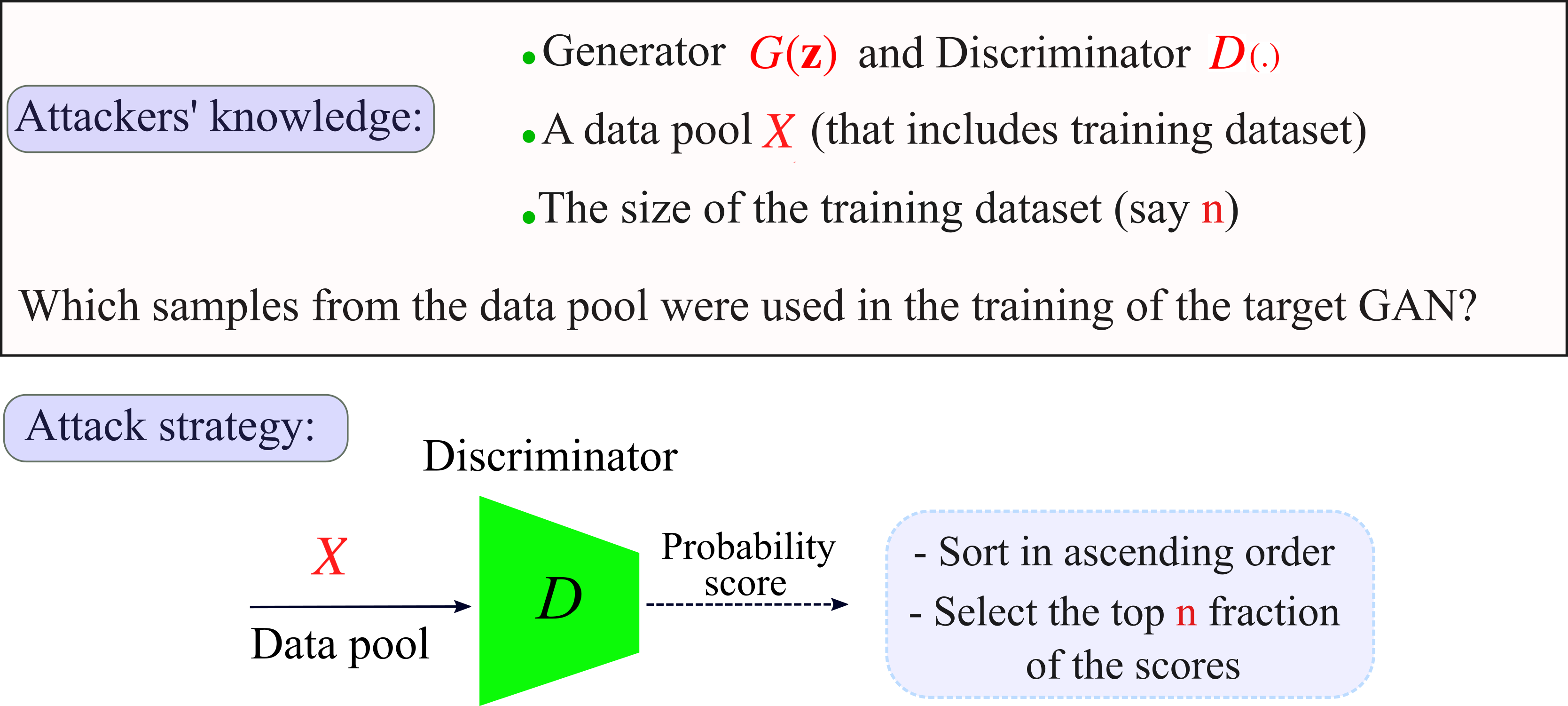}
    \caption{A membership inference attack strategy against GANs based on the target discriminator~\cite{hayes2019logan}. In its black-box format where the discriminator of the target GAN is not given, an auxiliary GAN based on the synthetic data points is trained, and then its discriminator is used instead.}
    \label{fig:haye_MIA}
\end{figure}

%\begin{equation}\label{MIA1}
%     M_{\text{GAN}}(x) = \mathbb{P}(x\in X_{tr}), \hspace{0.5cm} \forall x \in X.
%\end{equation}
%\begin{equation}\label{MIA1}
%     M_{\text{GAN}}(x) = \mathds{1}\{x\in X_{tr}\}, \hspace{0.5cm} \forall x \in X.
%\end{equation}
For discriminative models, the relation between the generalization gap and MIAs was studied in~\cite{yeom2018privacy}. Regarding GAN frameworks, this relation was proposed by Wu et.al.~\cite{wu2019generalization} where they theoretically showed how this generalization gap can be bounded for a GAN trained with a differentially private learning algorithm. In addition, this notion is asserted by other studies such as~\cite{sun2021adversarial,li2021membership,chen2021gan}. Incorporating this generalization gap in the attack mechanism is the main intuition beyond the membership inference attack proposed by Hayes et al.~\cite{hayes2019logan} which assumes a white-box scenario where the adversary not only can get queries from the discriminator but also has access to a substantial data pool and is aware of the size of the training dataset used in training the target generative model. As discussed before, this MIA has shown to be effective on GANs and has been considered widely for evaluating defense mechanisms~\cite{mukherjee2021privgan,chen2021gan}. More precisely, this attack is performed based on the discriminator response to the target data samples. When the generalization gap is large, it means the discriminator overfits to the training dataset and therefore it returns high values (probability values close to one) for the training data samples and small values for the hold-out and synthetic data samples. By knowing the size of the training dataset and having access to a pool of datasets, the attacker can incorporate the discriminator confidence on the pool data samples to infer the training data samples. Fig.~\ref{fig:haye_MIA} represents the details of this MIA strategy.

During this study, following the literature, we opt to evaluate our defense frameworks based on the proposed MIA in Fig.~\ref{fig:haye_MIA}. Additionally, in this work, we will utilize the TVD attack, as proposed in~\cite{mukherjee2021privgan}, to establish an upper limit for the attackers' performance in inferring the training samples.

\section{Proposed Frameworks} \label{sec:prop_fr}

\subsection{Measure of overfitting}

It was mentioned that the accuracy of the membership inference attacker is closely connected to the generalization gap. Thus, most of the defense mechanisms proposed in the literature, such as those based on differential privacy or more recent frameworks such as PrivGAN~\cite{mukherjee2021privgan} and PAR-GAN~\cite{chen2021gan}, were developed based on the idea of decreasing the generalization gap. In our work, we want to explore the relation between the generalization gap and the accuracy of membership inference attackers from another, more general, point of view. In fact, we show that although reducing the generalization gap can reduce the accuracy of MIAs, there might be other cases in which we can reduce the inference performance of the attacker without reducing the generalization gap.

\begin{figure*}[htbp!]
    \centering
    \includegraphics[width=0.95\linewidth]{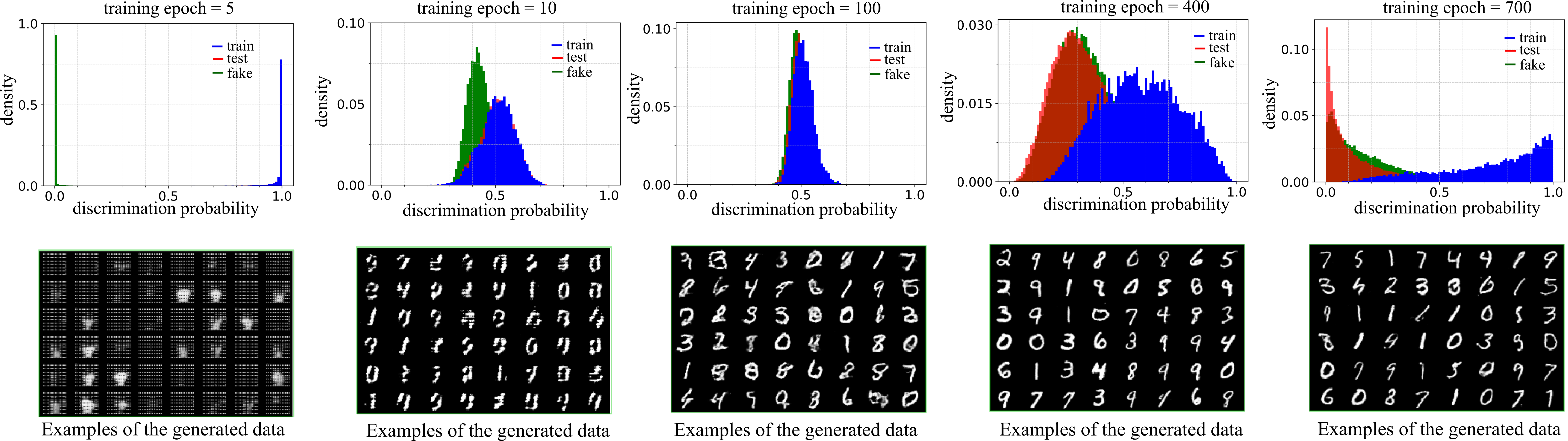}
    \caption{An example of the meaning of overfitting in the discriminator of a GAN in terms of the evolution of the discriminator response to training, test, and fake data samples over the training epoch.}
    \label{fig:overlap_GAN}
\end{figure*}

In a GAN framework, the discriminator is a binary classifier that aims to distinguish between the real data samples (training data samples) and the fake/synthetic sample points crafted by the generator. However, the notion of overfitting in the discriminator of GANs is different from that of a regular classifier. When the discriminator overfits on training samples, it usually returns significantly high values for the training data samples and small values for the other samples including fake and test samples. This point is discussed more empirically in Fig.~\ref{fig:overlap_GAN} where a GAN is trained on MNIST data and the density of the output of the discriminator is presented at different training epochs for the training, test, and fake data samples. From this figure, it can be seen that at the initial epochs, the discriminator returns high scores for the train and test data samples while very small values for the fake samples. Once the generator is able to generate real-looking samples, the discriminator responds the same for fake and real (train and test) data (see results of epoch 100). However, after several hundred epochs, the discriminator overfits on the training data and starts returning high values for train data samples and small values for the fake and test data points. It should be noted that the discriminator responses are almost the same for the fake and test data samples and the generator is able to generate real-looking samples. This can be observed in the generated samples in Fig.~\ref{fig:overlap_GAN}. It can clearly be seen from Fig.~\ref{fig:overlap_GAN} that overfitting in the discriminator increases the generalization gap. During the training process, although test data are not available, the generalization gap can be approximated from the discriminator responses to fake and train data samples. This generalization gap can be exploited by the attacker to infer the membership of data samples. However, it is not clear how the MIA accuracy and the generalization gap are related mathematically. In the following, we introduce the so-called Bhattacharyya coefficient, defined at the output of the discriminator, as a measure of overfitting and we discuss its relationship with the MIA performance. %and the discriminator response in terms of .%, as described below.%Below this point is framed from the GAN holder point of view.\\

%\begin{proposition}
Consider a GAN framework trained on a training data set $X_{TR}$. Let the score set $S$ be defined as follows:
\begin{equation*}
    S=\{s: s = D(x), \; x\in X\},
\end{equation*}
where $X$ is a publicly available data pool that includes $X_{TR}$. Consider an MIA that aims to infer the training set $X_{TR}$ using the score set $S$. Let $\omega_1$ and $\omega_0$ refer to the class of scores associated with the training and non-training data samples, respectively. The minimum error of any membership inference attacker modeled as a binary classifier $M^{d}_{\text{GAN}}: S \longrightarrow \{\omega_0,\omega_1\}$ is equal to the Bayes error defined below\cite{theodoridis2006pattern}:  
\begin{equation}
   P_e^{m} = \mathbb{P}(\boldsymbol{s} \in R_1,\omega_0) + \mathbb{P}(\boldsymbol{s}\in R_0,\omega_1),
\end{equation}
where $P_e^{m}$ is the error of membership inference attacker, $R_i\;, i=0,1$ is the score space in which the attacker decides in favor of $\omega_i$ and $\mathbb{P}(.,.)$ is the joint probability. 
As it is mentioned by Kailath~\cite{1089532}, considering $\pi_0$ and $\pi_1$ as the a priori probabilities, this error can be bounded as follows:
\begin{equation}\label{eq:BC_bound}
    \frac{1}{2} - \frac{1}{2}\sqrt{1-4\pi_0\pi_1\rho^2}\leq P_e^{m}\leq \sqrt{\pi_0\pi_1} \rho,
\end{equation}
where, considering likelihood functions $p(s|\omega_0)$ and $p(s|\omega_0)$, the Bhattacharyya coefficient $\rho$ is defined below~\cite{1089532}:
\begin{equation}\label{eq:BC_def}
    \rho \coloneqq \int \sqrt{p(s|\omega_0)p(s|\omega_1)}ds.
%    \rho \coloneqq \int \sqrt{p_{\mathbf{s}|\boldsymbol{\omega}}(s|\omega_0)p_{\mathbf{s}|\boldsymbol{\omega}}(s|\omega_1)}ds.
\end{equation}
%\end{proposition}

From the error bound~\eqref{eq:BC_bound} it can be seen that maximizing $\rho$ can be used to limit the performance of the membership inference attacker. In statistics, the Bhattacharyya coefficient is used as a separability measure between two classes/populations~\cite{theodoridis2006pattern}. Therefore, maximizing $\rho$ can be interpreted as maximizing the overlap (minimum separability) between score classes $\omega_0$ and $\omega_1$. This point is directly related to overfitting in the discriminator since $\omega_1$ (the class of score associated with the training samples) and $\omega_0$ (the class of score associated with the non-training samples) have maximum separability in the case of overfitting discriminator. This fact is illustrated in Fig.~\ref{fig:overlap_GAN} as discussed previously. As mentioned before, in this paper, we consider the Bhattacharyya coefficient $\rho$, instead of generalization gap $g_{\text{GAN}}^{D}$ in equation~\eqref{eq:gengap}, as a measure of overfitting. More specifically, a value of $\rho$ close to one (large overlap) is interpreted as a case with smaller overfitting, while a value of $\rho$ close to zero is associated with high overfitting in the GAN. 

In the following, we analyze in more detail the Bhattacharyya coefficient for the Gaussian scenario in order to gain more insight and discuss its relationship with the generalization gap.

\begin{proposition}~\cite{theodoridis2006pattern} Consider the Gaussian case where the densities $p(s|\omega_0)$ and $p(s|\omega_1)$ are, respectively, $\mathcal{N}\left(\mu_{0}, \sigma^2_{0}\right)$ and $\mathcal{N}\left(\mu_{1}, \sigma^2_{1}\right)$. In this case, the Bhattacharyya coefficient would be as follows:
\begin{equation}\label{eq:BC_GAUSSIAN}
    \rho = \exp{\left(-\frac{1}{4}\ln{\left[\frac{1}{4}\left(\frac{\sigma^2_{1}}{\sigma^2_{0}}+\frac{\sigma^2_{0}}{\sigma^2_{1}}+2\right)\right]} - \frac{1}{4}\left[\frac{(\mu_{0}-\mu_{1})^2}{\sigma^2_{0}+\sigma^2_{1}}\right]\right)}.
\end{equation}
\end{proposition}

From equation~\eqref{eq:BC_GAUSSIAN}, it can be seen that for fixed variances, reducing the differences between the mean values can increase $\rho$ (and thus reduce overfitting). This corresponds to the definition of the generalization gap. Nevertheless, equation~\eqref{eq:BC_GAUSSIAN} also illustrates that increasing the variances (even for fixed mean values) can increase the $\rho$. In other words, the Bhattacharyya coefficient is a more general indicator of overfitting compared with the generalization gap. From a GAN point of view, the Bhattacharyya coefficient between the score densities associated with training samples and fake samples can be calculated and used as a measure of overfitting in the discriminator. It is worth noting again that once the generator learns to generate real-looking samples, the discriminator responds almost equally to synthetic and test data samples (see Fig.~\ref{fig:overlap_GAN}). Thus, the Bhattacharyya coefficient associated with the training and synthetic samples can be used as an estimation of the Bhattacharyya coefficient associated with the training and test data samples.

In the next subsections, two new GAN frameworks will be presented as defense mechanisms against membership inference attacks and the Bhattacharyya coefficient defined in equation~\eqref{eq:BC_def} will be used to quantify the overfitting.

\subsection{Maximum Entropy GAN (MEGAN)}
%In the last section, the Bhattacharyya coefficient was introduced as a measure of overfitting in the discriminator and its relation with the MIA accuracy was discussed.
In this section, the first defense framework against membership inference attacks for GANs is presented. To this end, overfitting is studied in terms of the discriminator error (as a binary classifier) and we discuss how overfitting and discriminator error are inversely related. 

\begin{proposition}
In a GAN framework, the discriminator is a binary classifier that aims to measure the likelihood of each sample being real. Particularly, considering the binary random variable $\boldsymbol{re} \in \{r,f\}$, we can say:
\begin{equation}\label{eq:discrim}
    D(x) = p_{\boldsymbol{re}|\boldsymbol{x}}(r|x), \qquad \forall x\in X, \end{equation}
or equivalently, $p_{\boldsymbol{re}|\boldsymbol{x}}(f|x) = 1 - D(x)$ denotes the probability that $x$ is a fake sample. Let the regions $R_r$ and $R_f$, respectively, refer to the sample spaces in which we decide in favor of the sample being real and fake. Then, the Bayes error of the discriminator, as a binary classifier, can be written as follows:
\begin{align}\label{eq:dis_err}
   % P_e^{d} = & \mathbb{P}(x\in R_r,f) + \mathbb{P}(x\in R_f,r) \nonumber\\
%    & = \mathbb{P}(x\in R_r|f)p_{\mathbf{f}} + \mathbb{P}(x\in R_f|r)p_{\mathbf{r}} \\
 %   &= p_{\mathbf{f}}\int_{R_r}p_{\mathbf{x}|\mathbf{f}}dx + p_{\mathbf{r}}\int_{R_f}p_{\mathbf{x}|\mathbf{r}}dx\nonumber\\
  %  & \overset{\text{(i)}}{=} \int_{R_r}p_{\mathbf{f}|\mathbf{x}}p_{\mathbf{x}}dx + \int_{R_f}p_{\mathbf{r}|\mathbf{x}}p_{\mathbf{x}}dx \nonumber\\
   % &\overset{\text{(ii)}}{=} \int_{R_r}(1-D(x))p_{\mathbf{x}}dx + \int_{R_f}D(x)p_{\mathbf{x}}dx \nonumber\\
 %   &\overset{\text{(iii)}}{=}p_{\mathbf{r}} - \int_{R_r}(2D(x)-1)p_{\mathbf{x}}dx\nonumber\\
  %  &\overset{\text{(iv)}}{=}p_{\mathbf{f}} - \int_{R_f}(1-2D(x))p_{\mathbf{x}}dx\nonumber
    P_e^{d} & = \P(\boldsymbol{x}\in R_r,\boldsymbol{re} = f) + \mathbb{P}(\boldsymbol{r}\in R_f,\boldsymbol{re} = r) \nonumber\\
    & = \mathbb{P}(\boldsymbol{x} \in R_r| \boldsymbol{re} = f) p(f) + \mathbb{P}(\boldsymbol{x}\in R_f|\boldsymbol{re} = r) p(r) \\
    &= p(f) \int_{R_r} p(x|f) dx + p(r) \int_{R_f} p(x|r) dx\nonumber\\
    & \overset{\text{(i)}}{=} \int_{R_r} p(f|x) p(x) dx + \int_{R_f} p(r|x) p(x) dx \nonumber\\
    &\overset{\text{(ii)}}{=} \int_{R_r}(1-D(x))p(x) dx + \int_{R_f} D(x) p(x) dx \nonumber\\
    &\overset{\text{(iii)}}{=} p(r) - \int_{R_r}(2D(x)-1)p(x) dx\nonumber\\
    &\overset{\text{(iv)}}{=} p(f) - \int_{R_f}(1-2D(x)) p(x) dx. \nonumber
\end{align}
where (i) is due to the Bayes rule, (ii) is based on the equation~\eqref{eq:discrim}, and (iii) and (iv) are based on the following:
\begin{align*}
    p(r) & = \int_{R_r} D(x)p(x) dx + \int_{R_f}D(x) p(x) dx, \\
    p(f) & = \int_{R_f}(1-D(x))p(x) dx + \int_{R_r}(1-D(x))p(x) dx.
\end{align*}

When the discriminator overfits on the training data samples, $D(x)$ is generally larger for $x\in X_{tr} \subset R_r$ than for $x\notin X_{TR}$. Thus, from equation~\eqref{eq:dis_err}, both last and second last equalities, it can be seen that the discriminator error decreases as $D(x)$ increases in the region $x\in X_{tr}$. In other words, overfitting in the GAN discriminator is directly related to the error of the discriminator.  More specifically, a smaller error in the discriminator usually gives a larger overfitting on the training data samples. Therefore, to limit the performance of membership inference attacks, we need to control the minimum error of the discriminator of the GAN in order to avoid overfitting. This can be done by considering the so-called Fano's inequality from the field of information theory to obtain a bound on $P_e^d$.
\end{proposition}

%\francisco{I think we should add some more details to motivate the next proposition and explain the entropy notation.\textcolor{green}{Could you please do that? Thanks a lot.}}

\begin{proposition}
From Fano's inequality\cite{cover2006elements}, we have
\begin{equation}\label{eq:fano_dis}
    P_e^{d}\geq \frac{H(\mathbf{re}|\mathbf{x}) - 1}{\log 2} \approx \frac{\E\left[ H(D(\mathbf{x}))\right] - 1}{\log 2}.
\end{equation}
From ~\eqref{eq:fano_dis}, it can be seen that maximizing $\E[H(D(\mathbf{x}))]$ can avoid the discriminator error from being very small and thus can be used to control overfitting.%\\
\end{proposition}

Motivated by Proposition 4, we propose a novel framework to learn the distribution of training data and simultaneously reduce the accuracy of the membership inference attackers by decreasing overfitting through maximization of the lower bound in equation~\eqref{eq:fano_dis}. Similar to the GAN, this framework includes a generator and a discriminator but, unlike the GAN, the generator is trained to maximize the entropy of the discriminator outputs. More specifically, this framework named Maximum Entropy GAN (MEGAN) is developed by solving the following multi-objective optimization problem:
\begin{equation} \label{MEGAN_Opt}
    \text{MEGAN}:\;	\underset{D, G}{\text{max}}\;
    \left(h_1, h_2\right),
\end{equation}
\noindent with
\begin{align} \label{h1,h2}
h_1 =& \E\left[\log D(\mathbf{x_{tr}})\right] +\E\left[\log(1-D(G(\mathbf{z})))\right], \\
h_2 =&\E\left[H\Big(D(G(\mathbf{z}))\Big)\right], \\\nonumber
\end{align}
\noindent where $H(p) = -p\log p - (1-p)\log(1-p)$ is the entropy of a Bernoulli random variable with parameter $p$\cite{cover2006elements}. On the one hand, the first optimization problem in~\eqref{MEGAN_Opt} is exactly the discriminator optimization in the original GAN framework~\cite{goodfellow2014generative}. Therefore, for a fixed generator, following the proof proposed in~\cite{goodfellow2014generative} the optimum discriminator would be as follows:
\begin{equation}\label{Opt_discriminator}
    D^{*}(x) =\frac{p_{\mathbf{x_{tr}}}(x)}{p_{\mathbf{x_{tr}}}(x) + p_{\mathbf{x_{g}}}(x)}.
\end{equation}
On the other hand, since the binary entropy $H(p)$ is maximized for $p=0.5$, the solution for the second optimization problem in~\eqref{MEGAN_Opt} is $D(G(z))=0.5$. Thus, considering~\eqref{Opt_discriminator}, the MEGAN optimization problem~\eqref{MEGAN_Opt} converges to its optimum when $p_{\mathbf{x_{g}}} = p_{\mathbf{x_{tr}}}$. This means that, as in classical GANs, MEGAN can learn the underlying distribution of the training data points. However, in vanilla GANs, the generator learns the distribution through a minimax optimization by trying to fool the discriminator into confusing real and fake data samples, while the generator in MEGAN learns the distribution by maximizing the uncertainty of the discriminator. The experimental results will show how this can reduce the accuracy of MIAs. The training algorithm for MEGAN is explained in detail in Algorithm~\ref{AL_MEGAN}. It should be noted that, at each training iteration of the discriminator, the generator is generally trained for several iterations since, unlike the discriminator which is a simple binary classifier, the generator has a more difficult task. 

\begin{algorithm}
    %\scriptsize
    \footnotesize
    \algsetup{linenosize=\tiny}
	\caption{Maximum Entropy GAN (MEGAN): Batch size $B$ and number of steps to apply to the Generator $k$ are hyperparameters.}
	\label{AL_MEGAN}
	\begin{algorithmic}[1]
	\FOR {number of training iterations}
		\STATE Sample minibatch of $B$ noise samples $\{ z_{1},\dots, z_{B}\}$ from noise distribution $p_{\mathbf{z}}$.
		\STATE Sample minibatch of $B$ examples $\{ x_{1},\dots, x_{B}\}$ from training data set distribution $p_{\mathbf{x_{tr}}}$.
		\STATE Compute the gradient of $\mathcal{L}_{\mathcal{D}}(\theta_{d})$, approximated empirically for minibatch, with respect to $\theta_{d}$ and update $\theta_{d}$ by applying the Adam optimizer~\cite{kingma2014adam}.
		\vspace{5pt}
		
		$\mathcal{L}_{\mathcal{D}}(\theta_{d}) \coloneqq  - \frac{1}{n} \sum_{i=1}^{n}\left[ \log D(x_i) + \log(1-D(G(z_i)))\right].$
        
        \vspace{5pt}
		
		\FOR {$k$ steps}
		\STATE Sample minibatch of $B$ noise samples $\{ z_{1},\dots, z_{B}\}$ from noise distribution $p_{\mathbf{z}}$.
		\STATE Compute the gradient of $\mathcal{L}_{\mathcal{G}}(\theta_{g})$, approximated empirically for minibatch, with respect to $\theta_{g}$ and update $\theta_{g}$ by applying the Adam optimizer.
		
       $\mathcal{L}_{\mathcal{G}}(\theta_{g}) \coloneqq  \frac{1}{n}\sum_{i=1}^{n} \Big[D(G(z_i))\log D(G(z_i)) +$ \\
       \hspace{70pt}    $\left(1-D(G(z_i))\right)\log \left(1-D(G(z_i))\right)\Big].\nonumber$

       \vspace{5pt}
       \ENDFOR
		\vspace{5pt}
       
		\ENDFOR
	\end{algorithmic}
\end{algorithm}

\subsection{Mutual Information Minimization GAN (MIMGAN)}

In this section, we propose a more general framework motivated by the fact that synthetic samples might leak information regarding the training data points which can be used by membership inference attackers. To prevent such an information leakage, besides learning the training distribution, we train the generator to produce samples that give minimum information about the training data set $X_{tr} = \{x_{tr,i}\}_{i=1}^{n}$. By defining the joint random variable $\mathbf{x_{tr}^n} = (\mathbf{x_{tr,1}},\mathbf{x_{tr,2}},\dots,\mathbf{x_{tr,n}})$, the mutual information between the synthetic data and the training data points, i.e. $I(\mathbf{x_{tr}^n};\mathbf{x_{g}})$, is considered. Basically, minimizing this mutual information means that each generated sample should leak minimum information about the whole training dataset. However, minimizing this mutual information directly is generally very cumbersome. Thus, we look for a surrogate upper bound to optimize instead:
\begin{align}\label{MI_GAN_upp}
    I(\mathbf{x_{tr}^n};\mathbf{x_{g}})& = H(\mathbf{x_{tr}^n}) - H(\mathbf{x_{tr}^n}|\mathbf{x_{g}}) \nonumber \\ & \overset{\text{(i)}}{=}\sum_{i=1}^{n}H(\mathbf{x_{tr,i}}) - H(\mathbf{x_{tr}^n}|\mathbf{x_{g}}) \nonumber \\ & \overset{\text{(ii)}}{\leq}n\times H(\mathbf{x_{tr,j}}) - H(\mathbf{x_{tr,j}}|\mathbf{x_{g}}) \\ & = 
    (n-1)\times H(\mathbf{x_{tr}}) + I(\mathbf{x_{tr,j}};\mathbf{x_{g}}), \nonumber
\end{align}
where step (i) is due to the assumption that training data points are independent and identically distributed (i.i.d.) and (ii) is based on the fact that the joint entropy is larger than each marginal entropy, where $j\in [1,n]$ is selected randomly. Therefore, instead of working with the $I(\mathbf{x_{tr}^n};\mathbf{x_g})$, its upper bound in the equation~\eqref{MI_GAN_upp} can be minimized. Since for a fixed training data set the first term of the upper bound in the equation~\eqref{MI_GAN_upp} is a constant, the $I(\mathbf{x_{tr,j}};\mathbf{x_{g}})$ can be minimized. It should be emphasized that the index $j\in [1,n]$ here is selected randomly. This is along with the nature of GANs, where the outputs of the generator are not labeled and so generally there are no pre-defined pairs of samples in the format $(x_{tr}, x_{g})$. Therefore, in the process of training our GAN framework, for each generated sample, a random data point from the training dataset should be paired with it. From now on, for the sake of convenience, the term $I(\mathbf{x_{tr}};\mathbf{x_{g}})$ is used instead of $I(\mathbf{x_{tr,j}};\mathbf{x_{g}})$. This term can be added as a regularization term to the GAN loss in equation~\eqref{objective_vanillaGAN}. However, to simplify this regularization term, we consider an arbitrary conditional distribution $q_{\mathbf{x_{tr}}|\mathbf{x_{g}}}$ and note that:
\begin{align}\label{MI_GAN}
    I(\mathbf{x_{tr}};\mathbf{x_{g}})& = H(\mathbf{x_{tr}}) - H(\mathbf{x_{tr}}|\mathbf{x_{g}}) = H(\mathbf{x_{tr}}) + \nonumber \\
    & \mathbb{E}\left[\log q_{\mathbf{x_{tr}}|\mathbf{x_{g}}}\right] + \text{KL}\left(p_{\mathbf{x_{tr}}|\mathbf{x_{g}}}\|q_{\mathbf{x_{tr}}|\mathbf{x_{g}}}\right)\geq \nonumber\\
    &H(\mathbf{x_{tr}}) +\mathbb{E}\left[\log q_{\mathbf{x_{tr}}|\mathbf{x_{g}}}\right],
\end{align}
where $\text{KL}(.\|.)$ is the Kullback–Leibler divergence, a measure of how a probability distribution (the first term in KL function) is different from another probability distribution (the second term in KL function)~\cite{cover2006elements}, and the last inequality is due to the fact that the KL is non-negative. Since $\text{KL}\left(p_{\mathbf{x_{tr}}|\mathbf{x_{g}}}\|q_{\mathbf{x_{tr}}|\mathbf{x_{g}}}\right) = 0$ when $p_{\mathbf{x_{tr}}|\mathbf{x_{g}}} = q_{\mathbf{x_{tr}}|\mathbf{x_{g}}}$, the mutual information $I(\mathbf{x_{tr}};\mathbf{x_{g}})$, considering equation \eqref{MI_GAN}, can be written as follows:
\begin{align}\label{MI_GAN2}
    I(\mathbf{x_{tr}};\mathbf{x_{g}})& = H(\mathbf{x_{tr}}) + \underset{q_{\mathbf{x_{tr}}|\mathbf{x_{g}}}}{\text{max }}\mathbb{E}\left[\log q_{\mathbf{x_{tr}}|\mathbf{x_{g}}}\right],
\end{align}
where the expectation is with respect to the true distribution $p_{\mathbf{x_{tr}}|\mathbf{x_{g}}}$. Since the first term in equation~\eqref{MI_GAN2} is constant, the minimization of the mutual information $I(\mathbf{x_{tr}};\mathbf{x_{g}})$ would end up with a minimax problem between the generator and another network (named as adversary network). Adding the minimax game~\eqref{MI_GAN2} to the minimax formulation of the GAN in equation~\eqref{objective_vanillaGAN}, the total formulation of our proposed framework is as follows:
\begin{align}\label{objective_MIGAN}
    \underset{G}{\text{min}}\;\underset{D,\;A}{\text{max}}\; V(G,D,A)\; &= \E\left[\log D(\mathbf{x_{tr}})\right] +\nonumber \\
    &\E\left[\log(1-D(G(\mathbf{z})))\right] +\\
    &\lambda\times \E\left[\log q_{\mathbf{x_{tr}}|\mathbf{x_{g}}}\right],\nonumber
\end{align}
where $\lambda$ is the Lagrangian coefficient and $A(\mathbf{z};\theta_a)$ is the adversary network with parameter $\theta_a$. 

It should be noted that for the cases where $q_{\mathbf{x_{tr}}|\mathbf{x_{g}}}$ is the probability density function of a continuous random variable, it can be approximated by a conditional Gaussian distribution, i.e. $q_{\mathbf{x_{tr}}|\mathbf{x_{g}}}(x_{tr}|x_g) = \mathcal{N}\left(x_{tr}; \mu_{A}(x_g), \Sigma_{A}(x_g)\right)$ where $\mathcal{N}(x; \mu, \Sigma) = \det(2\pi \Sigma)^{-1/2} \exp\left[-\frac{1}{2}(x-\mu)^T\Sigma^{-1}(x-\mu)\right]$ is the probability distribution function of the Gaussian distribution~\cite{tripathy2019privacy}. Therefore, the GAN framework is updated by adding an adversary network that receives synthetic data at its input and outputs the vector of means $\mu_{A}$ and covariance matrix $\Sigma_{A}$ of the Gaussian distribution. More precisely, this adversary network aims to estimate the vector of means and the covariance matrix through maximizing $q_{\mathbf{x_{tr}}|\mathbf{x_{g}}}$ (maximum likelihood estimation). In this way, the generator learns to minimize $I(\mathbf{x_{tr}};\mathbf{x_{g}})$.
%\begin{figure}[htbp!]
%    \centering
%    \includegraphics[width=0.85\linewidth]{MIGAN_Framework.png}
%    \caption{General framework of the MIMGAN. An adversary deep neural network is considered.}
%    \label{fig:MIGAN}
%\end{figure}
%In addition, considering the minimax~\eqref{objective_MIGAN}, the loss function for each network of the Fig.~\ref{fig:MIGAN} would be as follows:
%\begin{align}
%    \mathcal{L}_{\mathcal{D}}(\theta_{d}) \coloneqq  - \frac{1}{n}&\sum_{i=1}^{n}\left[ \log D(\mathbf{x_{tr,i}}) + \log(1-D(G(\mathbf{z_{i}})))\right]\nonumber\\
%    \mathcal{L}_{\mathcal{A}}(\theta_{a}) \coloneqq  -\frac{1}{n}&\sum_{i=1}^{n} \log\mathcal{N}\left(\mathbf{x_{tr,i}}; \mu_{A}, \Sigma_{A}\right)\\
%    \mathcal{L}_{\mathcal{G}}(\theta_{g},\theta_{d},\theta_{a},\lambda) &\coloneqq  \frac{1}{n}\sum_{i=1}^{n} \Big[ -\log D(G(\mathbf{z_{i}})) + \nonumber\\
%    &\lambda\times\log\mathcal{N}\left(\mathbf{x_{tr,i}}; \mu_{A}, \Sigma_{A}\right)\Big]\nonumber
%\end{align}

The complete training algorithm for the proposed framework is explained in Algorithm~\ref{AL_MIGAN}. It should be noted that in the loss function of the generator, to get rid of the saturation, instead of minimizing $\E\left[\log(1-D(G(\mathbf{z})))\right]$ the term $\E\left[-\log(D(G(\mathbf{z})))\right]$ is minimized~\cite{goodfellow2014generative}.
\begin{algorithm}
    %\scriptsize
    \footnotesize
    \algsetup{linenosize=\tiny}
	\caption{MIMGAN: Privacy-preserving GAN based on minimum information leakage about the training data set. Batch size $B$ and the number of steps $k$ to apply to the adversary and discriminator networks are hyperparameters. The least expensive choice, i.e. $k=1$, is used in this study.}
	\label{AL_MIGAN}
	\begin{algorithmic}[1]
	\FOR {number of training iterations}
	    \FOR {$k$ steps}
		\STATE Sample minibatch of $B$ noise samples $\{ z_{1},\dots, z_{B}\}$ from noise distribution $p_{\mathbf{z}}$.
		\STATE Sample minibatch of $B$ examples $\{ x_{1},\dots, x_{B}\}$ from training data set distribution $p_{\mathbf{x_{tr}}}$.
		\STATE Compute the gradient of $\mathcal{L}_{\mathcal{D}}(\theta_{d})$, approximated empirically for minibatch, with respect to $\theta_{d}$ and update $\theta_{d}$ by applying the Adam optimizer~\cite{kingma2014adam}.
		\vspace{5pt}
		
		$\mathcal{L}_{\mathcal{D}}(\theta_{d}) \coloneqq  - \frac{1}{B} \sum_{i=1}^{B}\left[ \log D(x_i) + \log(1-D(G(z_{i})))\right]$
		
		\vspace{5pt}
		
		\STATE Compute the gradient of $\mathcal{L}_{\mathcal{A}}(\theta_{a})$, approximated empirically for minibatch, with respect to $\theta_{a}$ and update $\theta_{a}$ by applying the Adam optimizer.
		\vspace{5pt}
		
         $\mathcal{L}_{\mathcal{A}}(\theta_{a}) \coloneqq  -\frac{1}{B} \sum_{i=1}^{B} \log\mathcal{N}\left(x_i; \mu_{A}(G(z_{i})), \Sigma_{A}(G(z_{i}))\right)$
        
        \vspace{5pt}
        
		\ENDFOR
		
		\STATE Sample minibatch of $B$ noise samples $\{ z_{1},\dots, z_{B}\}$ from noise distribution $p_{\mathbf{z}}$.
		\STATE Sample minibatch of $B$ examples $\{ x_{1},\dots, x_{B}\}$ from training data set distribution $p_{\mathbf{x_{tr}}}$.
		\STATE Compute the gradient of $\mathcal{L}_{\mathcal{G}}(\theta_{g},\theta_{d},\theta_{a},\lambda)$, approximated empirically for minibatch, with respect to $\theta_{g}$ and update $\theta_{g}$ by applying the Adam optimizer.
		\vspace{5pt}

       $\mathcal{L}_{\mathcal{G}}(\theta_{g},\theta_{d},\theta_{a},\lambda) \coloneqq  \frac{1}{B}\sum_{i=1}^{B} \Big[ -\log D(G(z_{i})) +$ \\
       \hspace{70pt}    $\lambda\times\log\mathcal{N}\left(x_{i}; \mu_{A}(G(z_{i})), \Sigma_{A}(G(z_{i}))\right)\Big]\nonumber$
       
       \vspace{5pt}
       
		\ENDFOR
	\end{algorithmic}
\end{algorithm}

\section{Numerical Results and Discussion} \label{sec:results}
 In this study, four data sets that are commonly used in the literature are considered where the results of the MNIST and fashion-MNIST are presented here while the results for the Chest X-Ray Images (Pneumonia) dataset and Anime Faces dataset are shown in the Appendix. For all the experiments (except those that are explained explicitly) the models are trained on $10\%$ of the total data points and the other $90\%$ is used as the test or hold-out data points. Therefore, the accuracy of a random guessing strategy in inferring the training data points is $10\%$. The MIA strategy based on the discriminator, proposed in~\cite{hayes2019logan}, is used to assess the performance of the models. It is assumed that the attacker is aware of all the data points and knows that $10\%$ of that data was used to train the target GAN.
 
 \subsection{Effect of overfitting on the performance of MIA}
\begin{figure}[htbp!]
    \centering
    \includegraphics[width=0.80\linewidth]{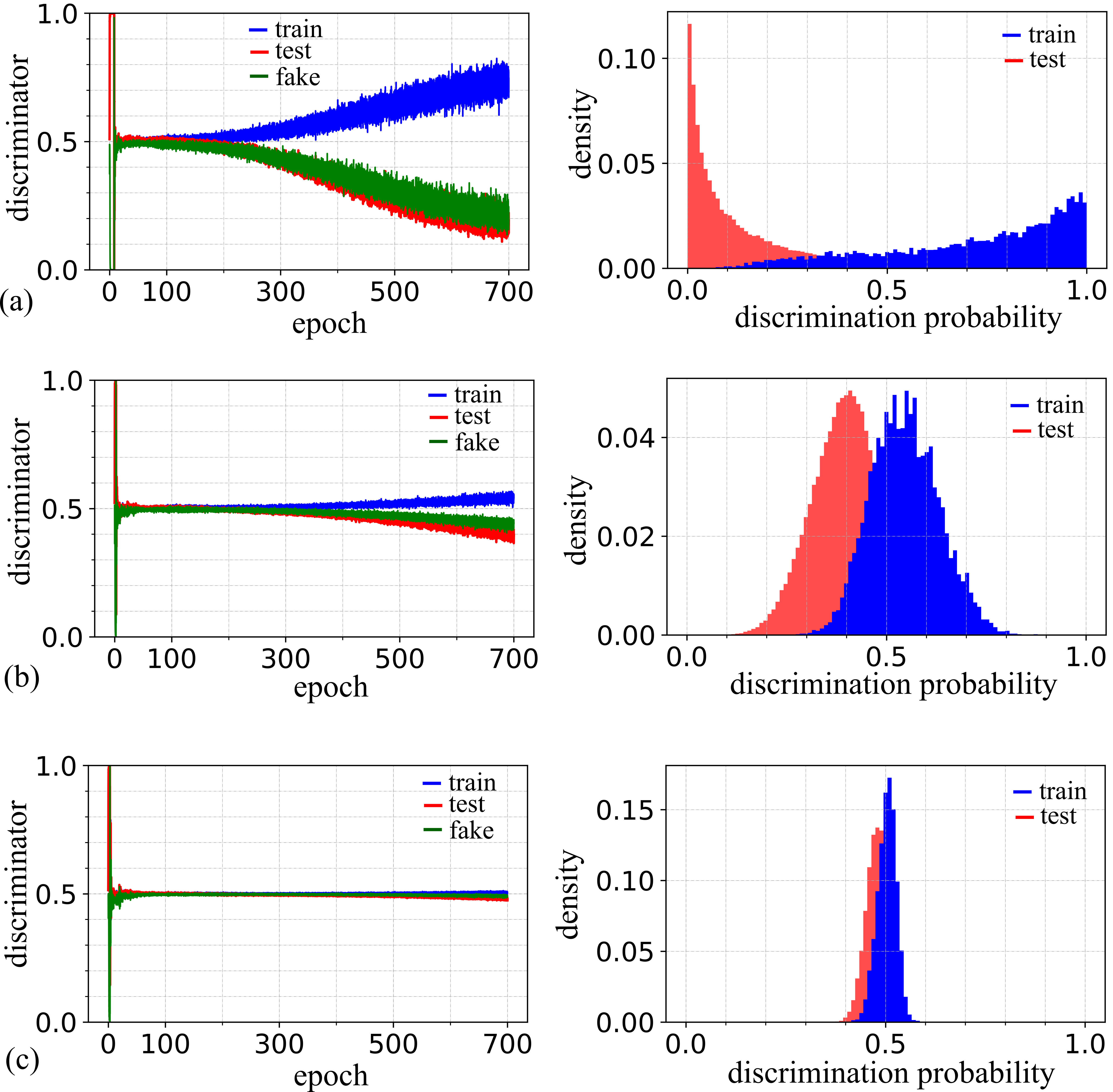}
    \caption{Examples of discriminator overfitting with (a) a highly overfitting case, (b) a case with moderate overfitting, and (c) a case with small overfitting. All three cases are trained on MNIST data set with the same train and test data samples but with different discriminator structures (see Appendix 1).}
    \label{fig:Overfitting_MNIST-res}
\end{figure}

As discussed previously, the membership inference attack proposed by Hayes et al.~\cite{hayes2019logan} relies on overfitting of the discriminator to the training data samples, i.e., the discriminator returns higher values for the training data than for the test (real unseen) data samples. In the following, we examine overfitting in relation to MIAs based on the MNIST dataset. To this end, different DCGAN frameworks are trained on the $10\%$ of the total MNIST data points. To check if the discriminator is overfitting to training data samples, the discriminator output for the train, test (hold-out), and the synthetic data points are monitored at each iteration and a gap between the discriminator output for the train and test data samples is considered as overfitting. The ideal case is when the discriminator returns $0.5$ for both train and test data. In addition, assigning the same value to test and synthetic data points by the discriminator is of interest since it means the generator is able to generate synthetic samples that match the real data samples. Fig.~\ref{fig:Overfitting_MNIST-res} presents the results of discriminator overfitting for three cases where GAN with very strong discriminator, strong discriminator, and mild discriminator is used, respectively. In this study, the overfitting in the discriminator is quantified by the Bhattacharyya coefficient defined in equation~\eqref{eq:BC_def} where the density of train and test data samples at the output of discriminator is used (see Fig.~\ref{fig:Overfitting_MNIST-res}). In addition, the memorization is measured using the $m_{\text{GAN}}^{G}$ defined in equation~\eqref{eq:memgan}. In Table~\ref{tab:overfitting}, the results of overfitting in discriminator (based on the generalization gap defined in equation~\eqref{eq:gengap} and the Bhattacharyya coefficient), memorization in generator (calculated based on $2000$ synthetic samples), and accuracy of MIA are presented for different GAN structures of Fig.~\ref{fig:Overfitting_MNIST-res}.

Considering Table~\ref{tab:overfitting}, from Fig.~\ref{fig:Overfitting_MNIST-res} (a) it can be seen that when the overfitting of the discriminator is significant, the distribution of the discriminator scores for training samples does not overlap perfectly with the one for test samples. This can be used by the attacker to take apart more confidently the training samples from the test samples. On the other hand, Figs.~\ref{fig:Overfitting_MNIST-res} (b) $\&$ (c) show that reducing overfitting in the discriminator can reduce the attacker's inference ability. 

\begin{table}[htbp]
	\centering
	\caption{Examining the discriminator overfitting and generator overfitting in relation to membership inference attack based on the MNIST data set. }% used for inference of households occupancy application
	\begin{adjustbox}{width=0.4\textwidth}
		\begin{tabular}{l c c c c c}
			\toprule
            \textbf{GAN structure*} & $g^{D}_{\text{GAN}}$ & $\rho$ & $m^{G}_{\text{GAN}}$ &MIA&\textbf{TVD Att.}\\
            \midrule[0.1pt]
            \textbf{\makecell{very strong\\ discriminator}}&0.55 &0.50& 1.18 & 60.37\% &0.75\\
            \midrule[0.1pt]
             \textbf{\makecell{strong \\discriminator}}&0.15 &0.66 & 0.98 & 54.36\%&0.59 \\
             \midrule
             \textbf{\makecell{mild \\discriminator}}&0.03 & 0.86& 0.96 & 31.91\% &0.35\\
             \midrule
			\bottomrule 
			\multicolumn{4}{l}{*see Appendix 1.}\\
		\end{tabular}
		%\centering
	\end{adjustbox}
	\label{tab:overfitting}
\end{table}

\begin{figure}[htbp!]
    \centering
    \includegraphics[width=0.85\linewidth]{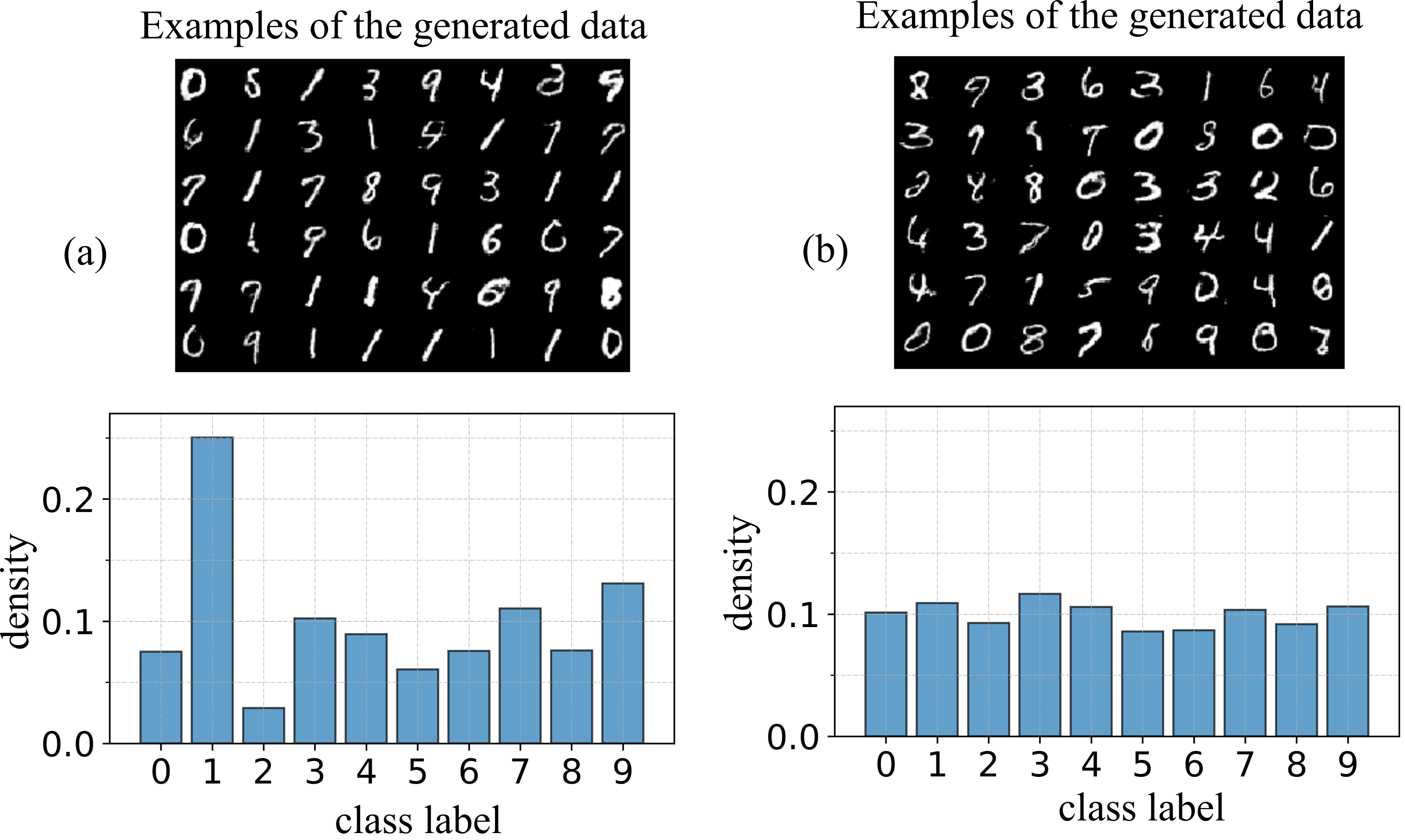}
    \caption{Comparing two GANs trained on MNIST data set wherein (a) the discriminator is highly overfitting to the training data samples - case (a) in Fig.~\ref{fig:Overfitting_MNIST-res}- and in (b) the overfitting of the discriminator is avoided as much as possible by reducing the gap between the discriminator outputs for train and test - case (c) in Fig.~\ref{fig:Overfitting_MNIST-res}-. The histograms present the class distributions based on a pre-trained classifier on MNIST data sets.}
    \label{fig:class_distri_Overfitting_MNIST-res}
\end{figure}

In addition to reducing the MIA accuracy, Fig.~\ref{fig:class_distri_Overfitting_MNIST-res} compares the case (a) in Fig.~\ref{fig:Overfitting_MNIST-res}, as an extreme case, with the case (c), as a mild case, in terms of the class distribution and samples quality.  
From this figure, it can be seen that although both cases can generate synthetic data samples of almost the same quality (same precision), the extreme case (in terms of the discriminator overfitting) tends to generate some classes more than others (small recall). 
%Moreover, higher entropy of the class distribution for the mild case, as the result of reducing overfitting in discriminator, is aligns with Inception Score (IS) in which one of the premises is that the marginal class label distribution should have high entropy. This emphasises on preventing discriminator overfitting.

It is worth noting that such a policy, i.e., simplifying the discriminator architecture, cannot completely prevent membership inference attackers from inferring membership labels. This can be seen from Table~\ref{tab:overfitting} where for the case with a mild discriminator, the white-box attack accuracy is still far from random guessing accuracy $10\%$. 

\subsection{Evaluation of the proposed frameworks against MIAs}
In this subsection, the experimental results of the proposed frameworks are presented and compared with the standard GAN. The model architectures and hyperparameters of each framework are listed in Appendix 2 and Appendix 3.

\begin{figure*}[htbp!]
    \centering
    \includegraphics[width=0.75\linewidth]{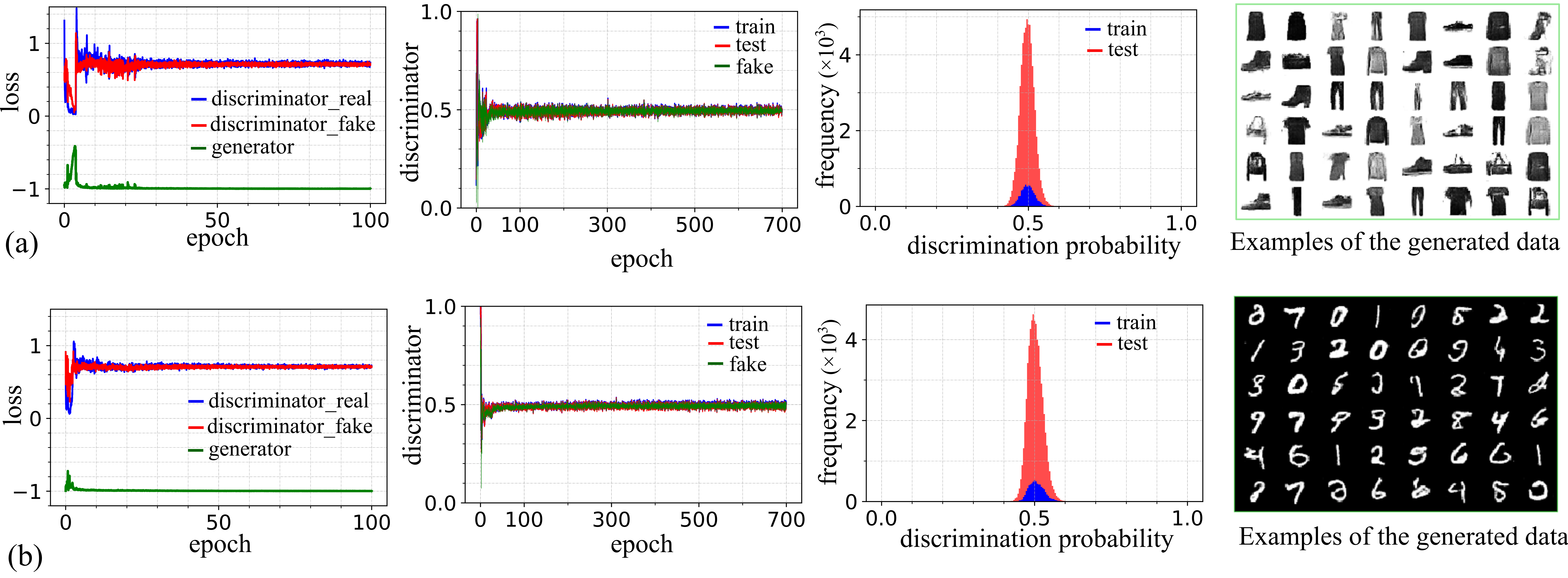}
    \caption{Examining the convergence in MEGAN based on the discriminator and generator loss functions, the output of discriminator per epoch, and the histogram of the output of discriminator used for white-box MIA on (a) fashion-MNIST and (b) MNIST data sets. For the sake of visibility, the loss functions are shown just for the first 100 epochs of the training process.}
    \label{fig:MEGAN_Conv}
\end{figure*}

First, we present the results of the MEGAN model trained on the MNIST and fashion-MNIST data sets. Fig.~\ref{fig:MEGAN_Conv} shows the convergence of the MEGAN in terms of the loss functions of the discriminator and generator. In addition, the discriminator output for the train, test, and synthetic data samples is represented. From Fig.~\ref{fig:MEGAN_Conv} it is clear that the MEGAN can converge and reach stability and is able to generate real-looking synthetic data samples. In addition, it can be seen that the density of the output of the discriminator for training and test data samples overlaps perfectly. Therefore, it can significantly reduce the accuracy of the membership inference attacker.

\begin{figure}[htbp!]
    \centering
    \includegraphics[width=0.95\linewidth]{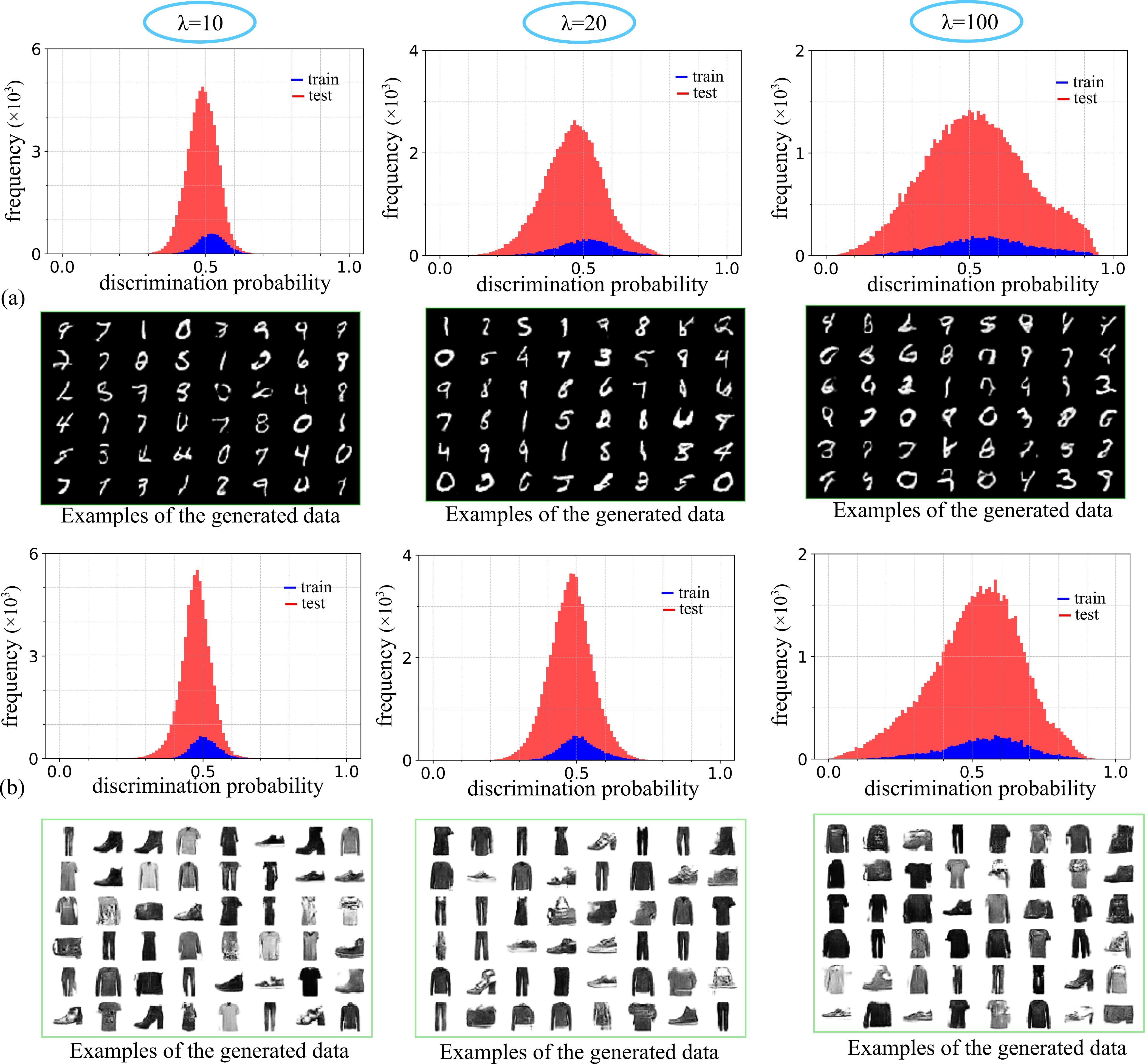}
    \caption{ Performance of the MIMGAN for different values of $\lambda$ applied to (a) MNIST and (b) fashion-MNIST data sets. For the sake of visibility, the histograms are not normalized.}
    \label{fig:MIMGAN_re}
\end{figure}

Regarding the MIMGAN, for different values of $\lambda$ (see equation~\eqref{objective_MIGAN}) the histogram of the discriminator output for the train and test data sets along with examples of the generated data samples is presented in Fig.~\ref{fig:MIMGAN_re}. From this figure, it can be seen that by increasing the value of $\lambda$ the amount of overlapping between the train and test density at the output of the discriminator increases. Thus, similar to MEGAN, we can expect MIMGAN to reduce the accuracy of the membership inference attackers. However, it should be noted that there is a clear difference between the MEGAN and MIMGAN approaches in reducing the accuracy of MIA. In MEGAN the overlapping between the discriminator output densities is done by making them distributed mainly around $0.5$ i.e., the optimum point of the discriminator, while the MIMGAN increases the variance of the discriminator output. To quantify this view, the overfitting of the discriminator and generator for GAN, MEGAN, and MIMGAN are examined in Table~\ref{tab:result_MEGAN} in terms of the accuracy of MIA and the overfitting parameters (i.e. generalization gap defined in equation~\eqref{eq:gengap} and the Bhattacharyya coefficient proposed in equation~\eqref{eq:BC_def}).  

Looking at Table~\ref{tab:result_MEGAN}, compared with GAN, it was expected that MEGAN would reduce the discriminator overfitting significantly. In addition, the generator in MEGAN has less overfitting than the one in GAN and so it is expected to reduce the accuracy of MIA extensively. The MIMGAN can also reduce the accuracy of MIA to discriminator where the overlapping parameter increases by increasing the value of $\lambda$ while the generalization gap is almost unchanged. The MIMGAN is a good example to show why Bhattacharyya coefficient $\rho$ should be used (instead of the generalization gap $g_{\text{GAN}}^D$) as the measure of overfitting in the discriminator. MIMGAN has also the best performance in terms of reducing memorization. Thus, we expect that for MIAs on the generator, MIMGAN would have the best performance.
\begin{table*}[htbp]
	\centering
	\caption{Comparison of the MEGAN and MIMGAN models with GAN in terms of overfitting, memorization, and MIA accuracy. }% used for inference of households occupancy application
	\begin{adjustbox}{width=0.67\textwidth}
		\begin{tabular}{l l l c c c c c}
			\toprule
            \textbf{Dataset} &\textbf{Model} & & \textbf{$g^{D}_{\text{GAN}}$} &\textbf{$\rho$}&\text{MIA*}& \textbf{$m^{G}_{\text{GAN}}$} &
            \textbf{TVD Att.}\\
            \midrule[0.1pt]
            \multirow{5}{*}{MNIST}&\textbf{GAN}& &0.03&0.86&31.91\% &0.96&0.35  \\ \cmidrule(lr){2-8}
                &\textbf{MEGAN}& &0.00&0.99&11.68\% &0.93 &0.06 \\ \cmidrule(lr){2-8}
                &\multirow{3}{*}{\textbf{MIMGAN}} &$\lambda=10$&0.03&0.95&21.94\% &0.94&0.25  \\ \cmidrule(lr){3-8}
                &&$\lambda=20$&0.04&0.97&16.73\% &0.89&0.18  \\ \cmidrule(lr){3-8}
                &&$\lambda=100$&0.04&0.99&11.30\% &0.81 &0.09 \\ 
             \midrule
             \multirow{5}{*}{fashion-MNIST}&\textbf{GAN}& &0.13&0.83 &33.41\%&0.93&0.39  \\ \cmidrule(lr){2-8}
                &\textbf{MEGAN}& &0.00&$\sim$1.00&11.01\% &0.86&0.02  \\ \cmidrule(lr){2-8}
                &\multirow{3}{*}{\textbf{MIMGAN}} &$\lambda=10$&0.020&0.94&21.71\% &0.90&0.24  \\ \cmidrule(lr){3-8}
                &&$\lambda=20$&0.03& 0.97&15.96\%&0.87&0.16 \\ \cmidrule(lr){3-8}
                &&$\lambda=100$&0.03& 0.99&11.47\%&0.82&0.08  \\
			\bottomrule
			\multicolumn{7}{l}{*Random attacker has MIA accuracy of 10\%. } \\
		\end{tabular}
		%\centering
	\end{adjustbox}
	\label{tab:result_MEGAN}
\end{table*}

\subsection{Precision-recall comparison with non-private GANs}\label{sec:prec-rec}

\begin{figure}[htbp!]
    \centering
    \includegraphics[width=0.99\linewidth]{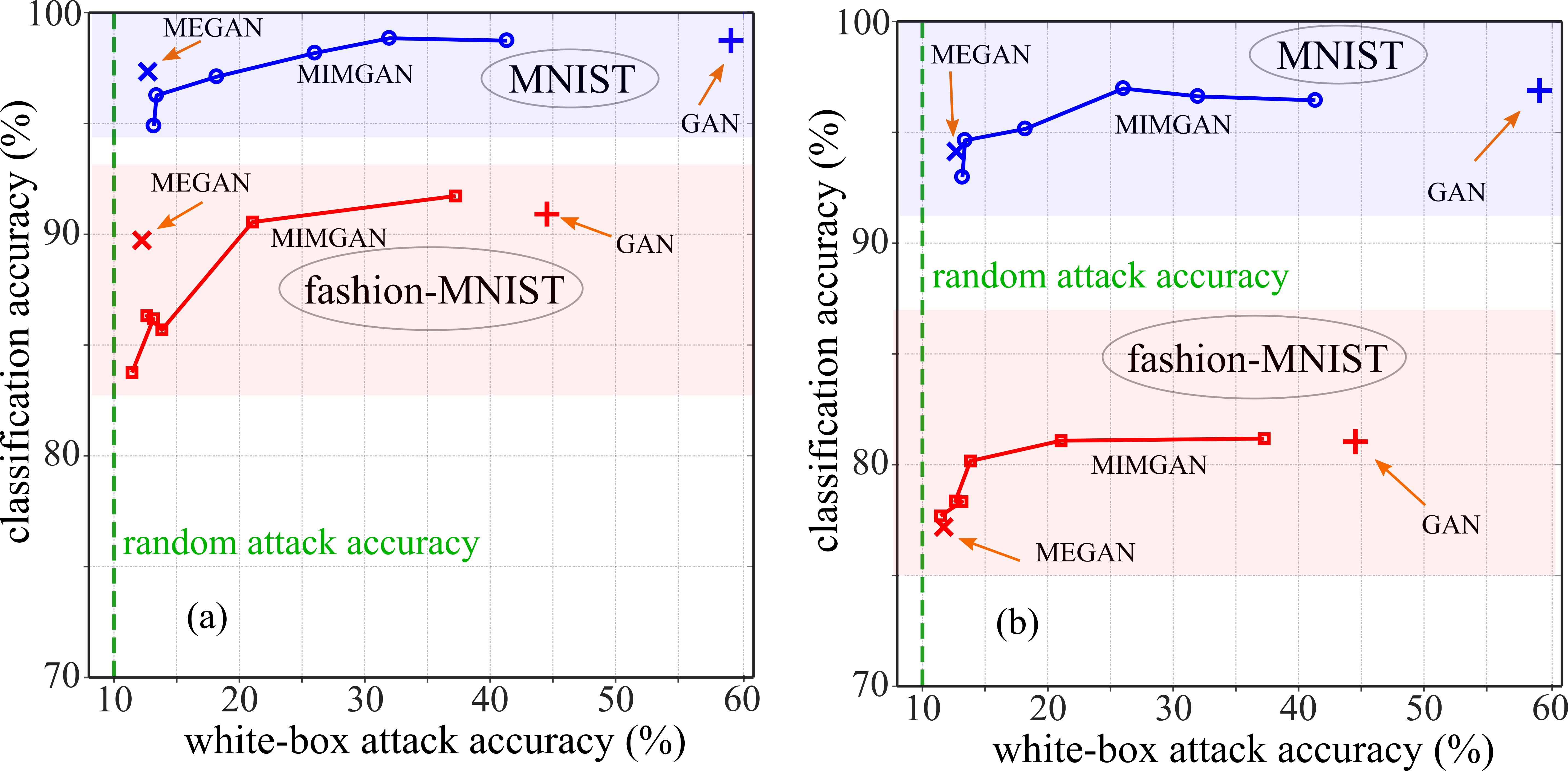}
    \caption{Evaluation of the GAN, MEGAN, and MIMGAN based on the (a) GAN-test and (b) GAN-train approaches versus the MIA accuracy.}
    \label{fig:utility_privacy}
\end{figure}

So far, our comparison has focused on assessing MEGAN and MIMGAN against the vanilla GAN in terms of their effectiveness in mitigating overfitting and their performance in countering MIAs. In terms of the quality (precision) and diversity (recall) of the generated data samples, although Figs.\ref{fig:MEGAN_Conv} $\&$ \ref{fig:MIMGAN_re} provide visual examples of the generated data points, these concepts can also be quantified through alternative metrics, such as GAN-test and GAN-train measures\cite{shmelkov2018good}. GAN-test represents the accuracy of a classifier trained on real data samples and evaluated on generated data points, akin to precision, where high accuracy signifies high-quality generated data samples. Conversely, GAN-train measures the accuracy of a classifier trained on generated samples and evaluated using real data points, similar to recall, where higher values indicate greater diversity among the generated samples. In this study, we employed the classifier outlined in Appendix 4 to compute GAN-test and GAN-train values for both MEGAN and MIMGAN, comparing them with the performance of the standard GAN. The results of this evaluation are illustrated in Fig.\ref{fig:utility_privacy}, alongside MIA accuracy. Examining this figure reveals that while both MEGAN and MIMGAN exhibit a slight reduction in precision and recall compared to the standard GAN, their substantial improvement in reducing MIA accuracy is evident. Furthermore, the figure highlights that in scenarios approaching extreme privacy (where MIA accuracy approximates randomness), MEGAN marginally outperforms MIMGAN in terms of precision and recall. However, MIMGAN offers flexibility by allowing control over the parameter $\lambda$, enabling users to reach their desired privacy-utility trade-off.

\subsection{Comparison with other private GANs}
In this section, our objective is to conduct a comparative analysis of our proposed models in comparison to other privacy-preserving GANs, specifically PrivGAN~\cite{mukherjee2021privgan} and DP-GAN~\cite{xie2018differentially}, across several key metrics. These metrics encompass MIA accuracy, the number of training parameters utilized within each framework, and GAN-test accuracy, as introduced in Section~\ref{sec:prec-rec}. It is important to emphasize that we employ a classifier with an identical structure, as detailed in Appendix 4, for all models. Additionally, the GAN structure remains consistent for each framework when applied to both datasets. Regarding DP-GAN, the same architecture as the vanilla GAN and MEGAN is used. Moreover, DP-GAN is implemented with differential privacy\footnote{https://github.com/tensorflow/privacy} with a parameter of $\delta=1e^{-4}$. Consequently, for each model, the initial step involves training the classifier using generated data samples, followed by its evaluation on a pre-determined test dataset. The results of this comprehensive comparison are presented in Table~\ref{tab:result_tot_comp}. The table unequivocally illustrates that while DP-GAN and PrivGAN exhibit the capacity to reduce MIA accuracy to a level approaching that of a random attacker, their performance in downstream utility is significantly compromised in comparison to MEGAN and MIMGAN, which are introduced in our work. More precisely, when we take into account both robustness against membership inference attacks and utility, as measured by GAN-test accuracy, our proposed models, MEGAN and MIMGAN, outperform Priv-GAN and DP-GAN. Notably, our MEGAN achieves this superiority while maintaining the same number of learning parameters as the standard GAN. In contrast, Priv-GAN has almost twice the number of parameters, making it significantly more computationally expensive. It is worth noting that this comparison is specifically conducted for the MNIST and fashion-MNIST datasets, primarily due to computational constraints.

\begin{table}[htbp]
	\centering
	\caption{Comparison of the MEGAN and MIMGAN models with other private GAN models. }% used for inference of households occupancy application
	\begin{adjustbox}{width=0.48\textwidth}
		\begin{tabular}{l l l c c c}
			\toprule
            \textbf{Dataset} &\textbf{Model} & &\textbf{MIA*}& \textbf{GAN-test Acc.} &
            \makecell{\textbf{$\#$ parameters}\\($\times 10^6$)}\\
            \midrule[0.1pt]
            \multirow{5}{*}{MNIST}&\textbf{GAN}& &59.20\% & 96.88\%& 4.69 \\ \cmidrule(lr){2-6}
                &\textbf{MEGAN}& &12.08\% & 94.16\%& 4.69\\ \cmidrule(lr){2-6}
                &\textbf{MIMGAN ($\lambda=100$)}& &13.01\% &92.97\% &9.65 \\ \cmidrule(lr){2-6}
                &\textbf{PrivGAN ($\lambda=10$, $N=2$)}& &12.18\% &77.51\% &9.49 \\ \cmidrule(lr){2-6}
                &\textbf{DP-GAN ($\delta=1e^{-4}$)}& &10.07\% &59.67\% & 4.69\\ 
             \midrule

             \multirow{5}{*}{fashion-MNIST}&\textbf{GAN}& &44.52\% &81.04\% &4.69  \\ \cmidrule(lr){2-6}
                &\textbf{MEGAN}& &12.24\% &77.24\% & 4.69\\ \cmidrule(lr){2-6}
                &\textbf{MIMGAN ($\lambda=100$)}& &11.44\% & 77.69\%& 9.65\\ \cmidrule(lr){2-6}
                &\textbf{PrivGAN ($\lambda=10$, $N=2$)}& &12.96\% & 67.07\%& 9.49\\ \cmidrule(lr){2-6}
                &\textbf{DP-GAN ($\delta=1e^{-4}$)}& &10.35\% & 56.23\%&4.69 \\ 
			\bottomrule
			\multicolumn{6}{l}{*Random attacker has MIA accuracy of 10\%. } \\
		\end{tabular}
		%\centering
	\end{adjustbox}
	\label{tab:result_tot_comp}
\end{table}

\section{Summary and Concluding Remarks} \label{sec:conclusion}

We have considered the problem of MIAs in GANs. First, we have revised the notion of overfitting in GANs and showed the limitations of the generalization gap. In particular, the Bhattacharyya coefficient between the distribution of the discriminator scores for training and non-training data points was introduced as a more complete measure of overfitting. The advantage of this coefficient is that it considers the shape of the distributions instead of only mean values. This was clarified with a Gaussian example in detail. Second, we proposed a new optimization framework for the GAN to mitigate the risks of membership inferences by maximizing the entropy of the discriminator scores for fake samples during training. This approach termed MEGAN was found to be quite effective in reducing the effectiveness of MIAs. Third, we consider another approach to try to mitigate the risks of MIAs by considering the leakage of information in fake samples. In this case, the GAN framework also was modified to include an additional network and a regularization term in the loss function to control the amount of information being leaked. The advantage of this scheme is that it provides a direct control, through the weight of the regularization term, over the amount of information leakage that is allowed. Thus, it is more flexible than the former approach. In all cases, there are trade-offs between the diversity and fidelity (quality) of the generated samples and the robustness against MIAs, as shown in Fig. \ref{fig:utility_privacy}. This is a topic worth of further research that may be considered in the future. 

\section*{Acknowledgment}
This work was supported by Hydro-Quebec, the Natural Sciences and Engineering Research Council of Canada, and McGill University in the framework of the NSERC/Hydro-Quebec Industrial Research Chair in Interactive Information Infrastructure for the Power Grid (IRCPJ406021-14). 

\bibliographystyle{ieeetr}
\bibliography{HREF}

\section*{Appendix 1: Model architectures for effect of discriminator overfitting on the performance of MIA}
The model architectures of the generators and discriminators used to produce Fig.~\ref{fig:Overfitting_MNIST-res} are presented along with the optimizer. It should be noted that the same generator was used for all three examples and an Adam optimizer with a learning rate of 0.0002 and $\beta = 0.5$ was used for optimization.\\

\noindent{\bf{Generator}}
\begin{enumerate}
    \item [-]Dense (units = 7$\times$7$\times$512, input size = 100)
    \item [-]LeakyReLU ($\alpha =0.2$)
    \item [-]Reshape ( target shape = (7,7,512)) 
    \item [-]Conv2DTranspose (filters = 128, kernel size = (5,5), strides = (2,2), padding = 'same')
    \item [-]LeakyReLU ($\alpha =0.2$)
    \item [-]Conv2DTranspose(filters = 128, kernel size = (5,5), strides = (2,2), padding = 'same')
    \item [-]LeakyReLU ($\alpha =0.2$)
    \item [-]Conv2D (filters = 1, kernel size =(5,5), activation = 'sigmoid',padding = 'same')
\end{enumerate}

\noindent{\bf{discriminator (a)}}
\begin{enumerate}
    \item [-]Conv2D (filters = 32, kernel size = (5,5), strides = (2,2), padding = 'same')
    \item [-]LeakyReLU ($\alpha =0.2$)
    \item [-]Conv2D (filters = 64, kernel size = (5,5), strides = (2,2), padding = 'same')
    \item [-]LeakyReLU ($\alpha =0.2$)
    \item [-]Conv2D(filters = 128, kernel size = (5,5), strides = (2,2), padding = 'same')
    \item [-]LeakyReLU ($\alpha =0.2$)
    \item [-]Flatten()
    \item [-]Dense(units = 1, activation = 'sigmoid')
\end{enumerate}

\noindent{\bf{discriminator (b)}}
\begin{enumerate}
    \item [-]Conv2D (filters = 64, kernel size = (5,5), strides = (2,2), padding = 'same')
    \item [-]LeakyReLU ($\alpha =0.2$)
    \item [-]Conv2D(filters = 128, kernel size = (5,5), strides = (2,2), padding = 'same')
    \item [-]LeakyReLU ($\alpha =0.2$)
    \item [-]Flatten()
    \item [-]Dense (units = 1, activation = 'sigmoid')
\end{enumerate}

\noindent{\bf{discriminator (c)}}
\begin{enumerate}
    \item [-]Conv2D (filters = 64, kernel size = (5,5), strides = (2,2), padding = 'same')
    \item [-]LeakyReLU ($\alpha =0.2$)
    \item [-]Conv2D(filters = 64, kernel size = (5,5), strides = (2,2), padding = 'same')
    \item [-]LeakyReLU ($\alpha =0.2$)
    \item [-]Flatten()
    \item [-]Dense (units = 1, activation = 'sigmoid')
\end{enumerate}

\section*{Appendix 2: Model architectures for the MEGAN}
The model architectures of the generators and discriminators used for MEGAN are presented along with the optimizer. The same architecture is used for both MNIST and fashion-MNIST datasets. The Adam optimizer with a learning rate of 0.0002 and $\beta = 0.5$ was used for optimization in Algorithm~\ref{AL_MEGAN}.\\

\noindent{\bf{Generator}}
\begin{enumerate}
    \item [-]Dense (units = 7$\times$7$\times$512, input size = 100)
    \item [-]LeakyReLU ($\alpha =0.2$)
    \item [-]Reshape ( target shape = (7,7,512)) 
    \item [-]Conv2DTranspose (filters = 128, kernel size = (5,5), strides = (2,2), padding = 'same')
    \item [-]LeakyReLU ($\alpha =0.2$)
    \item [-]Conv2DTranspose(filters = 128, kernel size = (5,5), strides = (2,2), padding = 'same')
    \item [-]LeakyReLU ($\alpha =0.2$)
    \item [-]Conv2D (filters = 1, kernel size =(5,5), activation = 'sigmoid',padding = 'same')
\end{enumerate}

\noindent{\bf{discriminator }}
\begin{enumerate}
    \item [-]Conv2D (filters = 32, kernel size = (5,5), strides = (2,2), padding = 'same')
    \item[-]BatchNormalization()
    \item [-]LeakyReLU ($\alpha =0.2$)
    \item [-]Conv2D(filters = 32, kernel size = (5,5), strides = (2,2), padding = 'same')
    \item[-]BatchNormalization()
    \item [-]LeakyReLU ($\alpha =0.2$)
    \item [-]Flatten()
    \item [-]Dense (units = 1, activation = 'sigmoid')
\end{enumerate}

\section*{Appendix 3: Model architectures for the MIMGAN}
The model architectures of the generator, discriminator, and Adversary used in MIMGAN to produce the results for MNIST and fashion-MNIST datasets are outlined here. For all cases, the same generator and Adversary architectures were used. In addition, an Adam optimizer with a learning rate of 0.0002 and $\beta = 0.5$ was used for optimization.\\

\noindent{\bf{Generator}}
\begin{enumerate}
    \item [-]Dense (units = 7$\times$7$\times$512, input size = 100)
    \item [-]LeakyReLU ($\alpha =0.2$)
    \item [-]Reshape ( target shape = (7,7,512)) 
    \item [-]Conv2DTranspose (filters = 128, kernel size = (5,5), strides = (2,2), padding = 'same')
    \item [-]LeakyReLU ($\alpha =0.2$)
    \item [-]Conv2DTranspose (filters = 128, kernel size = (5,5), strides = (2,2), padding = 'same')
    \item [-]LeakyReLU ($\alpha =0.2$)
    \item [-]Conv2D (filters = 1, kernel size = (5,5), activation = 'sigmoid', padding = 'same')
\end{enumerate}

\noindent{\bf{Adversary }}
\begin{enumerate}
    \item [-]Conv2D (filters = 64, kernel size = (3,3), strides = (2,2), padding = 'same')
    \item [-]LeakyReLU ($\alpha =0.2$)
    \item [-]Conv2D(filters = 64, kernel size = (3,3), strides = (2,2), padding = 'same')
    \item [-]LeakyReLU ($\alpha =0.2$)
    \item [-]Flatten()
    \item [-]Dense(units = 28$\times$28$\times$2)
\end{enumerate}

\noindent{\bf{discriminator (MNIST)}}
\begin{enumerate}
    \item [-]Conv2D (filters = 64, kernel size = (5,5), strides = (2,2), padding = 'same')
    \item [-]LeakyReLU ($\alpha =0.2$)
    \item [-]Conv2D (filters = 64, kernel size = (5,5), strides = (2,2), padding = 'same')
    \item [-]LeakyReLU ($\alpha =0.2$)
    \item [-]Flatten()
    \item [-]Dense (units = 1, activation = 'sigmoid')
\end{enumerate}

\noindent{\bf{discriminator (fashion-MNIST)}}
\begin{enumerate}
    \item [-]Conv2D (filters = 64, kernel size = (5,5), strides = (2,2), padding = 'same')
    \item [-]LeakyReLU ($\alpha =0.2$)
    \item [-]Conv2D(filters = 128, kernel size =(5,5), strides = (2,2), padding = 'same')
    \item [-]LeakyReLU ($\alpha =0.2$)
    \item [-]Flatten()
    \item [-]Dense (units = 1, activation = 'sigmoid')
\end{enumerate}

\section*{Appendix 4: Classifier architecture for GAN-test and GAN-train measures}

\begin{enumerate}
    \item [-]Conv2D (filters = 32, kernel size =(3,3), activation='relu')
    \item [-]MaxPooling2D(pool size = (2, 2))
    \item [-]Conv2D (filters = 64, kernel size =(3,3), activation='relu')
    \item [-]Conv2D (filters = 64, kernel size =(3,3), activation='relu')
    \item [-]MaxPooling2D (pool size = (2, 2))
    \item [-]Flatten()
    \item [-]Dense (units = 100, activation = 'relu')
    \item [-]Dense (units = 10, activation = 'softmax')
\end{enumerate}

\noindent SGD optimizer with a learning rate of 0.01 and momentum of 0.9 was used for optimization.

\section*{Appendix 5: More experimental results}
In this section, the experimental results of the proposed defense models applied to more datasets are proposed.

\begin{figure}[htbp!]
    \centering
    \includegraphics[width=0.85\linewidth]{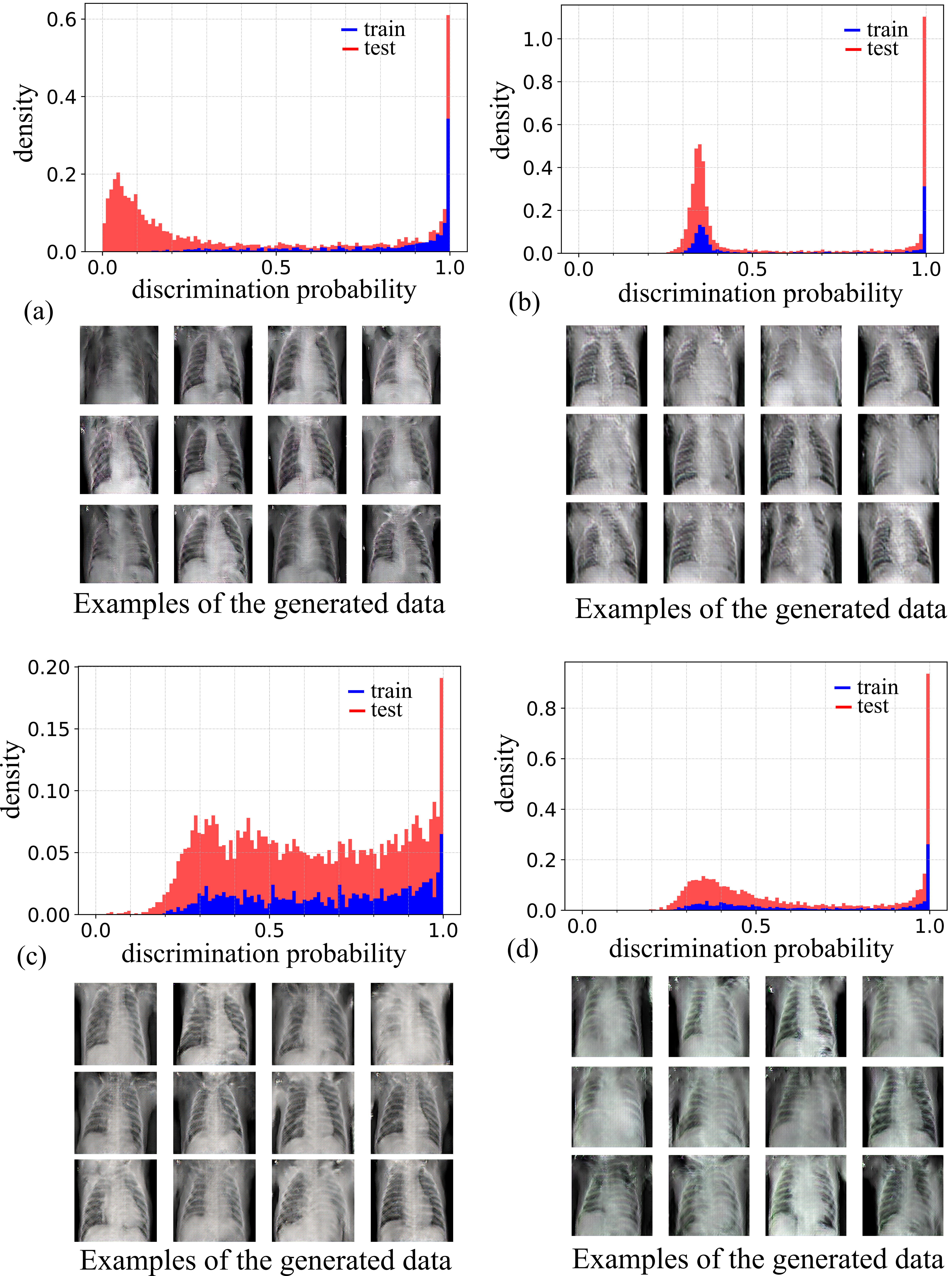}
    \caption{ Experimental results of (a) GAN, (b) MEGAN, (c) MIMGAN with $\lambda = 10$, and (d) MIMGAN with $\lambda = 20$ applied to the Chest X-Ray Images (Pneumonia) dataset~\cite{Kermany2018LabeledOC}. For this experiment, $80\%$ of the total data are considered as the test data set.}
    \label{fig:chestXray}
\end{figure}

\begin{table}[htbp]
	\centering
	\caption{Additional experimental results of the proposed models for the Chest X-Ray Images (Pneumonia) dataset.}
	\begin{adjustbox}{width=0.47\textwidth}
		\begin{tabular}{l l l c c c c c}
			\toprule
            \textbf{Dataset} &\textbf{Model} & & \textbf{$g^{D}_{gan}$} &\textbf{$\rho$}&\text{MIA*}& \textbf{$m^{G}_{gan}$}&\textbf{TVD Att.}\\
            \midrule[0.1pt]
            \multirow{3}{*}{\makecell{Chest X-Ray\\ Images (Pneumonia)}}&\textbf{GAN}& &0.39&0.74&36.39\% &1.01&0.49  \\ \cmidrule(lr){2-8}
                &\textbf{MEGAN}& &0.03&0.98&21.23\% &0.98 &0.06 \\ \cmidrule(lr){2-8}
                &\multirow{2}{*}{\textbf{MIMGAN}}
                &$\lambda=10$&0.05&0.97&24.49\% &0.96&0.09\\ \cmidrule(lr){3-8}
                &&$\lambda=20$&0.04&0.98&21.75\% &0.91&0.08  \\ 
			\bottomrule
			\multicolumn{7}{l}{*Random attacker has MIA accuracy of 20\%. } \\
		\end{tabular}
		%\centering
	\end{adjustbox}
	\label{tab:result_MEGAN_Xray}
\end{table}

Regarding the  Chest X-Ray Images (Pneumonia) dataset, the model architectures of the networks used in GAN, MIMGAN, and MEGAN are outlined here. For all cases, the kernels of each layer is initialized by a zero-mean random normal with $\sigma = 0.02$. Moreover, the Adam optimizer with a learning rate of 0.0002 and $\beta = 0.5$ was used.\\

\noindent{\bf{Generator}}
\begin{enumerate}
    \item [-]Dense (units = 8$\times$8$\times$128, input size = 100)
    \item [-]LeakyReLU ($\alpha =0.2$)
    \item [-]Reshape ( target shape = (8,8,128)) 
    \item [-]Conv2DTranspose (filters = 128, kernel size = (4,4), strides = (2,2), padding = 'same')
    \item [-]LeakyReLU ($\alpha =0.2$)
    \item [-]Conv2DTranspose (filters = 128, kernel size = (4,4), strides = (2,2), padding = 'same')
    \item [-]LeakyReLU ($\alpha =0.2$)
    \item [-]Conv2DTranspose (filters = 128, kernel size = (4,4), strides = (2,2), padding = 'same')
    \item [-]LeakyReLU ($\alpha =0.2$)
    \item [-]Conv2DTranspose (filters = 128, kernel size = (4,4), strides = (2,2), padding = 'same')
    \item [-]LeakyReLU ($\alpha =0.2$)
    \item [-]Conv2DTranspose (filters = 128, kernel size = (4,4), padding = 'same')
    \item [-]LeakyReLU ($\alpha =0.2$)
    \item [-]Conv2D (filters = 3, kernel size = (5,5), activation = 'sigmoid', padding = 'same')
\end{enumerate}

\noindent{\bf{Adversary (MIMGAN) }}
\begin{enumerate}
    \item [-]Conv2D (filters = 32, kernel size = (5,5), padding = 'same')
    \item [-]LeakyReLU ($\alpha =0.2$)
    \item [-]Conv2D(filters = 32, kernel size = (5,5), strides = (2,2), padding = 'same')
    \item [-]LeakyReLU ($\alpha =0.2$)
    \item [-]Conv2D(filters = 32, kernel size = (5,5), strides = (2,2), padding = 'same')
    \item [-]LeakyReLU ($\alpha =0.2$)
    \item [-]Conv2D(filters = 32, kernel size = (5,5), strides = (2,2), padding = 'same')
    \item [-]LeakyReLU ($\alpha =0.2$)
    \item [-]Conv2D(filters = 32, kernel size = (5,5), strides = (2,2), padding = 'same')
    \item [-]LeakyReLU ($\alpha =0.2$)
    \item [-]Flatten()
    \item [-]Dense(units = 128$\times$128$\times$3$\times$2)
\end{enumerate}

\noindent{\bf{discriminator (MIMGAN, GAN)}}
\begin{enumerate}
    \item [-]Conv2D (filters = 64, kernel size = (5,5), padding = 'same')
    \item [-]LeakyReLU ($\alpha =0.2$)
    \item [-]Conv2D (filters = 64, kernel size = (5,5), strides = (2,2), padding = 'same')
    \item [-]LeakyReLU ($\alpha =0.2$)
    \item [-]Conv2D (filters = 64, kernel size = (5,5), strides = (2,2), padding = 'same')
    \item [-]LeakyReLU ($\alpha =0.2$)
    \item [-]Conv2D (filters = 32, kernel size = (5,5), strides = (2,2), padding = 'same')
    \item [-]LeakyReLU ($\alpha =0.2$)
    \item [-]Conv2D (filters = 32, kernel size = (5,5), strides = (2,2), padding = 'same')
    \item [-]LeakyReLU ($\alpha =0.2$)
    \item [-]Conv2D (filters = 32, kernel size = (5,5), strides = (2,2), padding = 'same')
    \item [-]LeakyReLU ($\alpha =0.2$)
    \item [-]Flatten()
    \item [-] Dropout(0.5)
    \item [-]Dense (units = 1, activation = 'sigmoid')
\end{enumerate}

\noindent{\bf{discriminator (MEGAN)}}
\begin{enumerate}
    \item [-]Conv2D (filters = 32, kernel size = (5,5), padding = 'same')
    \item [-]LeakyReLU ($\alpha =0.2$)
    \item [-]Conv2D (filters = 32, kernel size = (5,5), strides = (2,2), padding = 'same')
    \item [-]LeakyReLU ($\alpha =0.2$)
    \item [-]Conv2D (filters = 32, kernel size = (5,5), strides = (2,2), padding = 'same')
    \item [-]LeakyReLU ($\alpha =0.2$)
    \item [-]Conv2D (filters = 32, kernel size = (5,5), strides = (2,2), padding = 'same')
    \item [-]LeakyReLU ($\alpha =0.2$)
    \item [-]Conv2D (filters = 32, kernel size = (5,5), strides = (2,2), padding = 'same')
    \item [-]LeakyReLU ($\alpha =0.2$)
    \item [-]Conv2D (filters = 32, kernel size = (5,5), strides = (2,2), padding = 'same')
    \item [-]LeakyReLU ($\alpha =0.2$)
    \item [-]Flatten()
    \item [-] Dropout(0.5)
    \item [-]Dense (units = 1, activation = 'sigmoid')
\end{enumerate}

\begin{figure}[htbp!]
    \centering
    \includegraphics[width=0.85\linewidth]{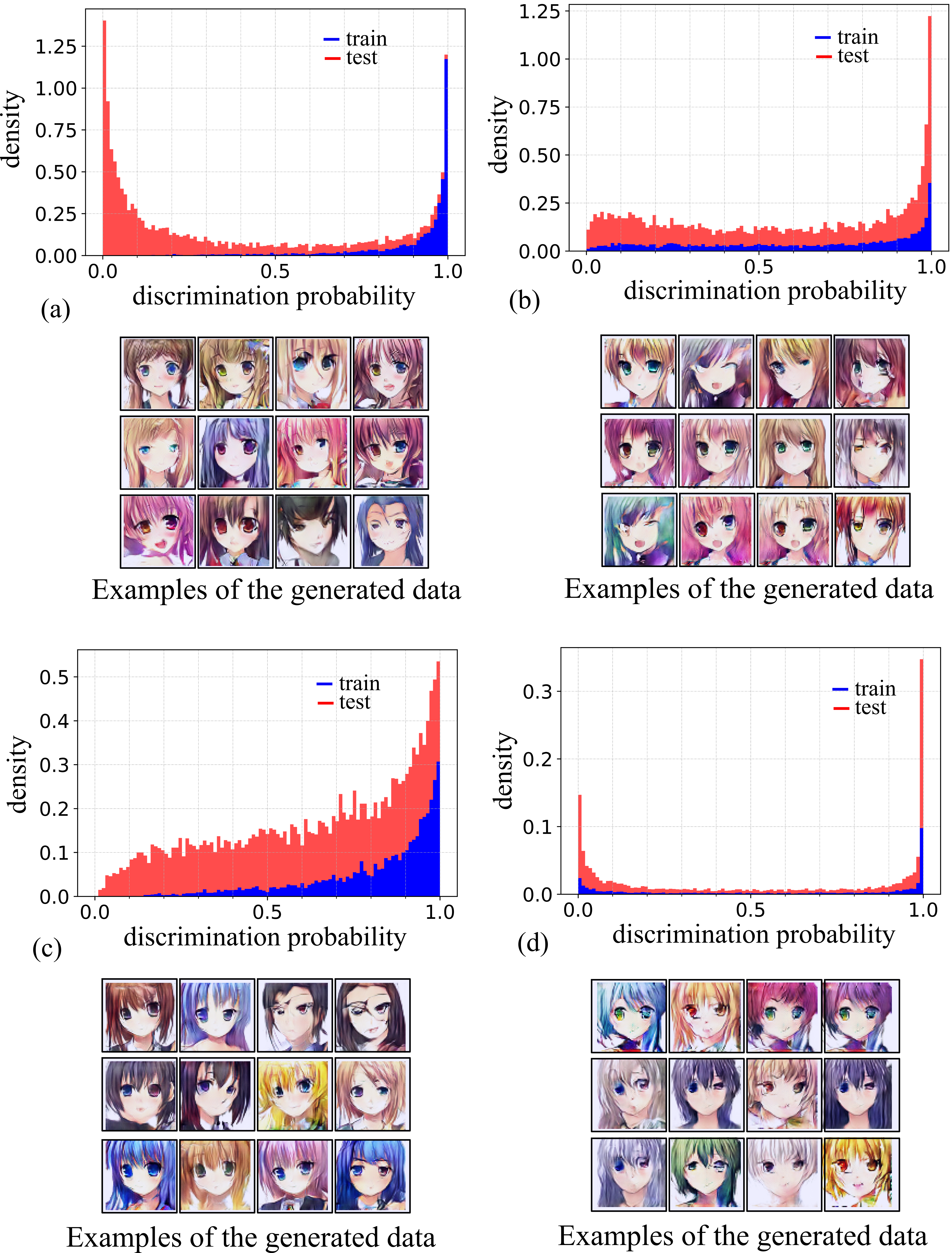}
    \caption{ Experimental results of (a) GAN, (b) MEGAN, (c) MIMGAN with $\lambda = 20$, and (d) MIMGAN with $\lambda = 100$ applied to the Anime Faces dataset. For this experiment, $80\%$ of the total data are considered as the test data set.}
    \label{fig:Animface}
\end{figure}

\begin{table}[htbp]
	\centering
	\caption{Additional experimental results of the proposed models for the Anime Faces dataset.}
	\begin{adjustbox}{width=0.47\textwidth}
		\begin{tabular}{l l l c c c c c}
			\toprule
            \textbf{Dataset} &\textbf{Model} & & \textbf{$g^{D}_{gan}$} &\textbf{$\rho$}&\text{MIA*}& \textbf{$m^{G}_{gan}$}&\textbf{TVD Att.}\\
            \midrule[0.1pt]
            \multirow{3}{*}{\makecell{Anime Faces}}&\textbf{GAN}& &0.48&0.70&48.53\% &0.762&0.58  \\ \cmidrule(lr){2-8}
                &\textbf{MEGAN}& &0.06&0.99&22.32\% &0.777 &0.08 \\ \cmidrule(lr){2-8}
                &\multirow{2}{*}{\textbf{MIMGAN}}
                &$\lambda=20$&0.17&0.93&33.77\% &0.757&0.27\\ \cmidrule(lr){3-8}
                &&$\lambda=100$&0.06&0.99&21.80\% &0.743&0.07  \\ 
			\bottomrule
			\multicolumn{7}{l}{*Random attacker has MIA accuracy of 20\%. } \\
		\end{tabular}
		%\centering
	\end{adjustbox}
	\label{tab:result_MEGAN_Anim}
\end{table}

Regarding the  Anime Faces dataset, the model architectures of the networks used in GAN, MIMGAN, and MEGAN are outlined here where in all the cases the Adam optimizer with a learning rate of 0.0002 and $\beta = 0.5$ was used.\\

\noindent{\bf{Generator}}
\begin{enumerate}
    \item [-]Dense (units = 4$\times$4$\times$512, input size = 100)
    \item [-]LeakyReLU ($\alpha =0.2$)
    \item [-]Reshape ( target shape = (4,4,512)) 
    \item [-]Conv2DTranspose (filters = 512, kernel size = (4,4), strides = (2,2), padding = 'same')
    \item [-] BatchNormalization(momentum=0.9, epsilon=0.0001)
    \item [-]LeakyReLU ($\alpha =0.2$)
    \item [-]Conv2DTranspose (filters = 256, kernel size = (4,4), strides = (2,2), padding = 'same')
    \item [-] BatchNormalization(momentum=0.9, epsilon=0.0001)
    \item [-]LeakyReLU ($\alpha =0.2$)
    \item [-]Conv2DTranspose (filters = 128, kernel size = (4,4), strides = (2,2), padding = 'same')
    \item [-] BatchNormalization(momentum=0.9, epsilon=0.0001)
    \item [-]LeakyReLU ($\alpha =0.2$)
    \item [-]Conv2DTranspose (filters = 64, kernel size = (4,4), strides = (2,2), padding = 'same')
    \item [-] BatchNormalization(momentum=0.9, epsilon=0.0001)
    \item [-]LeakyReLU ($\alpha =0.2$)
    \item [-]Conv2D (filters = 3, kernel size = (5,5), activation = 'tanh', padding = 'same')
\end{enumerate}

\noindent{\bf{Adversary (MIMGAN) }}
\begin{enumerate}
    \item [-]Conv2D (filters = 32, kernel size = (5,5), padding = 'same')
    \item [-]LeakyReLU ($\alpha =0.2$)
    \item [-]Conv2D(filters = 32, kernel size = (5,5), strides = (2,2), padding = 'same')
    \item [-]LeakyReLU ($\alpha =0.2$)
    \item [-]Conv2D(filters = 32, kernel size = (5,5), strides = (2,2), padding = 'same')
    \item [-]LeakyReLU ($\alpha =0.2$)
    \item [-]Conv2D(filters = 32, kernel size = (5,5), strides = (2,2), padding = 'same')
    \item [-]LeakyReLU ($\alpha =0.2$)
    \item [-]Conv2D(filters = 32, kernel size = (5,5), strides = (2,2), padding = 'same')
    \item [-]LeakyReLU ($\alpha =0.2$)
    \item [-]Flatten()
    \item [-]Dense(units = 64$\times$64$\times$3$\times$2)
\end{enumerate}

\noindent{\bf{discriminator (MIMGAN, GAN)}}
\begin{enumerate}
    \item [-]Conv2D (filters = 64, kernel size = (3,3), padding = 'same')
    \item [-]LeakyReLU ($\alpha =0.2$)
    \item [-] BatchNormalization()
    \item [-]Conv2D (filters = 64, kernel size = (3,3), padding = 'same')
    \item [-]LeakyReLU ($\alpha =0.2$)
    \item [-] BatchNormalization()
    \item [-] MaxPooling2D(pool size=(3,3))
    \item [-] Dropout(0.2)
    \item [-]Conv2D (filters = 64, kernel size = (3,3), padding = 'same')
    \item [-]LeakyReLU ($\alpha =0.2$)
    \item [-] BatchNormalization()
    \item [-]Conv2D (filters = 64, kernel size = (3,3), padding = 'same')
    \item [-]LeakyReLU ($\alpha =0.2$)
    \item [-] BatchNormalization()
    \item [-] MaxPooling2D(pool size=(3,3))
    \item [-] Dropout(0.3)
    \item [-]Flatten()
    \item [-] Dense(64)
    \item [-]LeakyReLU ($\alpha =0.2$)
    \item [-] Dense(64)
    \item [-]LeakyReLU ($\alpha =0.2$)
    \item [-]Dense (units = 1, activation = 'sigmoid')
\end{enumerate}

\noindent{\bf{discriminator (MEGAN)}}
\begin{enumerate}
    \item [-]Conv2D (filters = 64, kernel size = (3,3), padding = 'same')
    \item [-]LeakyReLU ($\alpha =0.2$)
    \item [-] BatchNormalization()
    \item [-]Conv2D (filters = 64, kernel size = (3,3), padding = 'same')
    \item [-]LeakyReLU ($\alpha =0.2$)
    \item [-] BatchNormalization()
    \item [-] MaxPooling2D(pool size=(3,3))
    \item [-] Dropout(0.2)
    \item [-]Conv2D (filters = 64, kernel size = (3,3), padding = 'same')
    \item [-]LeakyReLU ($\alpha =0.2$)
    \item [-] BatchNormalization()
    \item [-]Conv2D (filters = 64, kernel size = (3,3), padding = 'same')
    \item [-]LeakyReLU ($\alpha =0.2$)
    \item [-] BatchNormalization()
    \item [-] MaxPooling2D(pool size=(3,3))
    \item [-] Dropout(0.3)
    \item [-]Flatten()
    \item [-] Dense(32)
    \item [-]LeakyReLU ($\alpha =0.2$)
    \item [-] Dense(32)
    \item [-]LeakyReLU ($\alpha =0.2$)
    \item [-]Dense (units = 1, activation = 'sigmoid')
\end{enumerate}

\end{document}